%% file: iclr2027_conference.tex
\documentclass{article}
\usepackage{iclr2027_conference,times}

\input{math_commands.tex}

\usepackage{hyperref}
\usepackage{url}
\usepackage{amsmath}
\usepackage{amssymb}
\usepackage{graphicx}
\usepackage{booktabs}
\usepackage{multirow}
\usepackage{makecell}
\usepackage{float}
\usepackage{caption}

\title{Feature Puzzles: Single-Image Supervision for Multi-View 3D-Aware Segmentation}

\author{Jiahao Ma$^{1,3}$ \quad Qiang Zhang$^{2}$ \quad Miaomiao Liu$^{1}$ \quad David Ahmedt-Aristizabal$^{3}$ \quad Chuong Nguyen$^{3}$ \\
$^{1}$Australian National University \quad
$^{2}$University of Science and Technology of China \quad
$^{3}$CSIRO, Australia \\
}

\iclrfinalcopy
\begin{document}

\maketitle

% arxiv version
\thispagestyle{plain}   
\fancyhead{}            
\chead{\small Feature Puzzles: Single-Image Supervision for Multi-View 3D-Aware Segmentation}

\begin{figure}[H]
\centering
\includegraphics[width=0.95\textwidth]{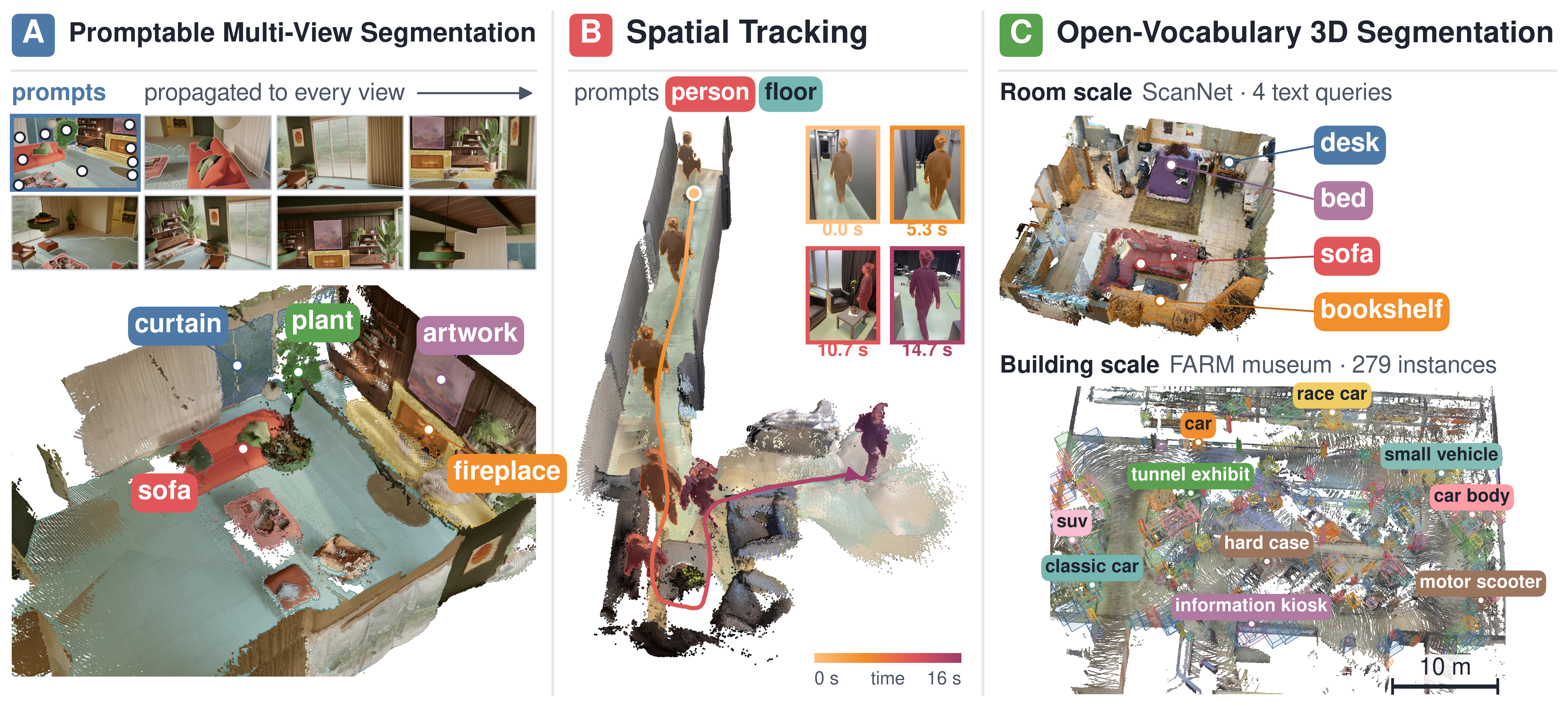}
\caption{\textbf{Feature Puzzles} learns multi-view-consistent segmentation from \textit{single images}, extending feed-forward reconstruction to (A) promptable segmentation that propagates to all views in one pass, (B) spatial tracking, and (C) open-vocabulary 3D segmentation.}
\label{fig:cover}
\end{figure}

\begin{abstract}
Humans can easily recognize the same objects across different viewpoints of a 3D scene, whereas teaching models such multi-view-consistent perception typically requires videos or calibrated multi-view captures, which are considerably more expensive and less diverse than large-scale monocular data.~In this paper, we introduce \textbf{Feature Puzzles}, a novel framework for learning multi-view-consistent segmentation using only single-image semantic segmentation supervision. Specifically, we propose a simple yet effective pipeline that constructs multi-view-consistent segmentation pseudo-labels through image cropping and geometric transformations. We further introduce a lightweight, plug-and-play network that leverages the multi-view-consistent image features learned by existing pretrained 3D foundation models and the generated pseudo-labels to perform multi-view-consistent segmentation. Importantly, our flexible framework supports various types of prompts provided from one or multiple views and can be readily extended to open-vocabulary 3D segmentation. Under this single-image supervision, our feed-forward model achieves the best mIoU on NVOS, remains competitive on SPIn-NeRF and ScanNet++ against baselines using per-scene optimization or in-domain multi-view training, and demonstrates zero-shot transfer to open-vocabulary 3D segmentation. Visual comparisons are available on our project page: \url{https://featurepuzzles.github.io/}.
\end{abstract}

%--------------------------------------------------------------------------------------
%--------------------------------------------------------------------------------------
%--------------------------------------------------------------------------------------
\section{Introduction}
\vspace{-0.3cm}

Spatial intelligence requires not only recovering geometric structure but also recognizing and reasoning about scene content consistently across viewpoints~\citep{yang2025thinking}. Multi-view-consistent perception, particularly multi-view-consistent segmentation, is essential for building coherent representations of the physical world and provides a foundation for downstream tasks such as robotic interaction~\citep{qu2025spatialvla,meshmimic}, embodied navigation~\citep{zhang2024vision,robotpan}, and open-vocabulary 3D understanding~\citep{peng2023openscene}.

Recent advances in feed-forward 3D reconstruction~\citep{wang2025vggt,wang2026pi3,lin2025depthanything3,wang2025spann3r,wang2025cut3r,lan2026stream3r} have substantially improved the efficiency of recovering 3D geometry while enabling the processing of long sequences. Building on these advances, IGGT~\citep{li2026iggt} and IGGT4D~\citep{zou2026iggt4d} extend feed-forward reconstruction models with instance-level understanding of static and dynamic scenes, while EmbodiedSplat~\citep{embodiedsplat2026} enables online 3D reconstruction and semantic understanding. Despite their effectiveness, these approaches typically rely on multi-view-specific resources, such as posed RGB-D scans, curated cross-view instance datasets, or multi-view distillation and self-supervision signals, which are considerably more difficult and costly to acquire than single-image semantic supervision.

This reveals a key supervision gap: pretrained multi-view backbones already provide strong geometric priors, but consistent cross-view semantics still depend on scarce multi-view annotations. In contrast, single-image segmentation data is abundant but lacks cross-view correspondence. We therefore ask whether single-image semantic supervision alone can supply this missing semantic layer without real multi-view annotations. We emphasize the scope of this question. We do not claim that 3D or multi-view consistency itself can be learned from monocular images alone; the geometric prior comes from the pretrained backbone, and our contribution is a supervision framework that adapts this representation to multi-view-consistent semantics.

To address this gap, we propose \textbf{Feature Puzzles}, a training framework that 
turns large-scale single-image segmentation data into synthetic multi-view 
supervision. It decomposes each image into overlapping crops and treats them as synthetic views, avoiding the need for multiple observations of the same scene.
%Since all crops originate from the same image coordinate system, their cross-view correspondences are exact,
For these shared-parent crop views, correspondence is exact in the parent-image coordinate system, 
enabling randomly sampled 2D masks and dense visual features to be transformed consistently 
across views without additional multi-view semantic annotation.

Building on this supervision, we introduce a lightweight adaptation scheme for feed-forward 3D reconstruction models. We fuse intermediate multi-scale features and adapt only late backbone blocks with low-rank adapters, which can be disabled at inference to recover the original depth and camera outputs. This equips existing backbones with multi-view-consistent perception without redesigning the reconstruction pipeline. We instantiate Feature Puzzles for two complementary tasks: 
\emph{promptable multi-view segmentation}, which accepts prompts from one or more views and predicts masks jointly across all views in a single forward pass, 
and \emph{open-vocabulary 3D segmentation}, which distills DINOv3 features to learn dense, multi-view-consistent representations for 3D recognition. As shown in Figure~\ref{fig:cover}, Feature Puzzles extends one reconstruction backbone to promptable segmentation, spatial tracking, and open-vocabulary 3D understanding with minimal architectural changes.
% \david{The claim that the method “largely preserves” geometric reconstruction performance is still unsupported by an explicit before/after geometry experiment. The supported claim is that disabling the adapters recovers the original reconstruction outputs. Unless a geometry-preservation experiment exists, this stronger claim should not remain.}

% \jiahao{We added an explicit before/after geometry experiment to support the claim that the method "largely preserves" geometric reconstruction performance.}

We organize our evaluation around a single hypothesis: 
\emph{shared-parent single-image supervision with exact synthetic-view correspondence can teach multi-view-consistent semantics to a pretrained 3D geometry representation}. 
Each experiment tests a consequence of this hypothesis. Promptable segmentation tests whether object identity transfers across genuine camera views. Instance spatial tracking tests whether this correspondence survives spatial and temporal displacement. Driving scenes test whether it transfers beyond indoor distributions. Open-vocabulary 3D segmentation tests whether the same principle extends from masks to dense semantic feature fields. 
For task-specific training, promptable segmentation uses only single-image SA-1B mask supervision, while open-vocabulary 3D segmentation uses single-image DINOv3 feature distillation. Neither task uses task-specific multi-view or video annotations. Under this supervision, our model achieves the best mIoU on NVOS in a single forward pass, remains competitive on SPIn-NeRF and ScanNet++, and demonstrates transfer to out-of-distribution driving scenes. Ablations further support the contributions of multi-layer features, backbone adaptation, and cross-view processing.

% \david{as mentioned in the Abstract, “trained only on SA-1B” should be scoped to task-specific Feature Puzzles training, since the system inherits pretrained VGGT-Omega and DINOv3}

% \jiahao{We  clarify that the SA-1B supervision is only used for task-specific training, and the system inherits pretrained VGGT-Omega and DINOv3.}

%--------------------------------------------------------------------------------------
%--------------------------------------------------------------------------------------
%--------------------------------------------------------------------------------------
\section{Related Work}
\vspace{-0.2cm}

\noindent\textbf{Feed-forward 3D reconstruction.}
Reconstruction has moved from per-scene optimization~\citep{schonberger2016sfm,schonberger2016mvs} to feed-forward prediction: VGGT~\citep{wang2025vggt}, $\pi^3$~\citep{wang2026pi3}, MapAnything~\citep{keetha2026mapanything}, and Depth Anything 3~\citep{lin2025depthanything3} regress cameras, depth, and point maps in a single pass, while  Spann3R~\citep{wang2025spann3r}, CUT3R~\citep{wang2025cut3r}, LingBot-Map~\citep{lingbotmap}, and STream3R~\citep{lan2026stream3r} extend this to streaming or causal reconstruction. Closest to our data construction, Puzzles~\citep{ma2025puzzles} synthesizes multi-view training data from a single image or clip, but only for geometric pointmap regression, without transferring semantic masks or features or defining a cross-view perception objective. These reconstruction models primarily focus on geometric structure rather than task-specific semantics. We build on a pretrained multi-view backbone and ask how single-image supervision can add a consistent semantic layer.

\noindent\textbf{Feed-forward semantic 3D scene understanding.}
Per-scene methods lift 2D vision--language features into a NeRF or 3DGS field~\citep{kerr2023lerf,zhou2024feature3dgs,qin2024langsplat,shi2024legaussians,wu2024opengaussian,occamslgs2025,drsplat2025,instancegaussian2025,hashpoint} and must re-optimize for every scene. Feed-forward alternatives predict semantics together with geometry in one pass. LSM~\citep{fan2024lsm}, Uni3R~\citep{sun2026uni3r}, SAB3R~\citep{chen2025sab3r}, and Ov3R~\citep{gong2026ov3r} distill CLIP- or DINO-style features into a reconstruction network, while EmbodiedSplat~\citep{embodiedsplat2026} builds a semantic 3DGS online from RGB(-D) streams. 
For instance understanding, PanSt3R~\citep{zust2025panst3r} adds a panoptic head to frozen MUSt3R and DINOv2 features, EPS3D~\citep{zhu2026eps3d} distills LSeg and SAM into a VGGT-based model, IGGT~\citep{li2026iggt}, IGGT4D~\citep{zou2026iggt4d} and S4VY~\citep{zhang2026s4vy} learn cross-view instance embeddings, and SceneDINO~\citep{jevtic2025scenedino} predicts a 3D DINO feature field from a single image. 
Most rely on multi-view scene data---including posed scans, curated cross-view instance datasets, or video---during semantic training. 
In contrast, Feature Puzzles uses no task-specific real multi-view or video supervision, while inheriting its geometric prior from a pretrained multi-view backbone. Its promptable branch uses single-image SA-1B masks, and its open-vocabulary branch a frozen DINOv3 teacher; cross-view supervision is created by shared-parent pull-back. Optional in-domain fine-tuning is reported separately from zero-shot evaluation.

% \david{The distinction must be scoped to task-specific semantic training/adaptation, because Feature Puzzles itself inherits a pretrained multi-view geometric backbone. "Uses no multi-view or video data of any kind" is therefore not defensible for the complete system. Also, the comparison should not imply that every competing method depends specifically on dense manually labelled cross-view data; some use distilled or self-supervised multi-view signals. The strong novelty claim is that Feature Puzzles does not require task-specific real multi-view/video semantic supervision or associated cross-view data construction.}

\noindent\textbf{Promptable and cross-view segmentation.}
SAM~\citep{kirillov2023sam} and SAM2~\citep{ravi2024sam2} provide promptable segmentation for images and video but do not explicitly model 3D cross-view consistency, so masks can drift across views of a static scene. 
NVOS~\citep{ren2022nvos}, SPIn-NeRF~\citep{mirzaei2023spinnerf}, SA3D~\citep{cen2023sa3d}, SAGA~\citep{cen2023saga, cen2025sa3dgs}, DCHM~\citep{dchm} and OmniSeg3D~\citep{ying2024omniseg3d} restore consistency by distilling SAM masks into a per-scene field, at a cost of minutes to hours per scene. MASA~\citep{li2024masa} learns 2D instance association from augmented static images, while MV-SAM~\citep{jeong2026mvsam} trains on SA-1B but relies on a frozen $\pi^3$ pointmap predictor for cross-view consistency at inference. The closest work is VGGT-Segmentor~\citep{gao2026vggtsegmentor}, which also builds on VGGT and offers a single-image self-supervised variant trained on SA-1B. It formulates cross-view segmentation as \emph{pairwise} source--target matching guided by explicit VGGT point tracks.
Because targets are processed independently, larger unordered view sets  require additional pair selection, association, or tracking.
Feature Puzzles instead trains and reasons jointly over shared-parent $N$-view sequences, predicting all views in one forward pass without pairwise matching, point tracking, or predefined view ordering.

%--------------------------------------------------------------------------------------
%--------------------------------------------------------------------------------------
%--------------------------------------------------------------------------------------
\section{Method}
\vspace{-0.2cm}

\subsection{Overview}

We propose Feature Puzzles, which turns a single image into overlapping synthetic views with exactly aligned supervision (Section~\ref{sec:puzzles}). A pretrained VGGT-Omega backbone encodes these views (Section~\ref{sec:backbone}) and feeds two lightweight heads for promptable multi-view segmentation (Section~\ref{sec:promptable}) and open-vocabulary 3D features (Section~\ref{sec:openvocab}).

\subsection{Single-Image Multi-Crop Puzzles}
\label{sec:puzzles}

\begin{figure}[t]
    \centering
    \includegraphics[width=\linewidth]{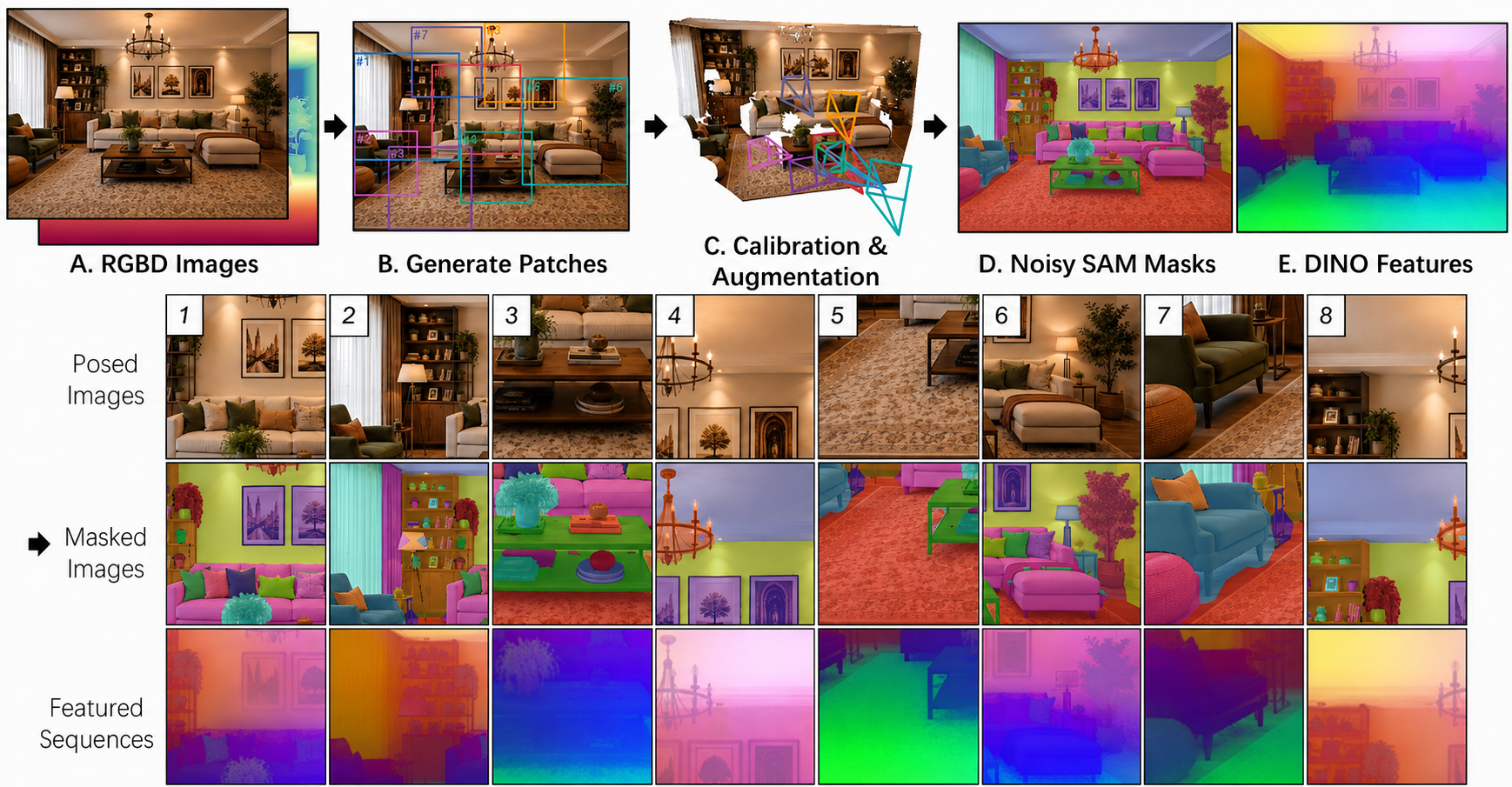}
    \vspace{-0.5cm}
    \caption{\textbf{Feature Puzzles pipeline.}
    From a single RGB(-D) parent image~(\textbf{A}), we sample overlapping crops as a puzzle~(\textbf{B}), optionally calibrate their cameras and apply 3D augmentation~(\textbf{C}), and pull back SAM masks~(\textbf{D}) and DINO features~(\textbf{E}) into every crop~(bottom).
    % Shared parent coordinates give cross-view correspondences for the affine crop views without videos or multi-view captures; optional 3D augmentation instead uses depth projection to synthesize viewpoint changes.
    }
    \label{fig:feature_puzzles_pipeline}
    \vspace{-0.2cm}
\end{figure}

% Figure~\ref{fig:feature_puzzles_pipeline} summarizes how Feature Puzzles turns one annotated image into a pseudo-view training sample. The idea is simple but effective: we call the original image the \emph{parent image}, which carries not only RGB but also instance masks, dense teacher features, and semantic classes; through Feature Puzzles we crop it into several overlapping \emph{sub-images}, each treated as a view. Because the sub-images originate from the same parent, their labels in overlapping regions agree by construction, yielding multi-view-consistent supervision without videos or multi-view captures.
Figure~\ref{fig:feature_puzzles_pipeline} shows how Feature Puzzles converts one annotated image into a pseudo-view sample. We call the original image the \emph{parent image}, containing RGB, instance masks, dense teacher features, and semantic labels. It is cropped into overlapping \emph{sub-images}, each treated as a synthetic view. Since all sub-images share the same parent, overlapping regions inherit exactly aligned supervision by construction, enabling multi-view-consistent training without real multi-view or video data.

% \david{Please distinguish exact synthetic-view correspondence from real camera-view correspondence.}

% \david{a bit unclear, you establish exact correspondence between transformed observations of the same parent pixels. That is not the same as exact correspondence between views of a 3D scene under camera movement.}
% \jiahao{Puzzle views are generated in two modes, and the text documents only the first.
% 1.Affine puzzle crops. Overlapping boxes in parent coordinates, related to the parent by a closed-form affine map πₙ. These carry no parallax and no occlusion, and — as you correctly observe — their correspondence is exact only in the sense of relating transformed observations of the same parent pixels.
% 2.Projected puzzle views. We also retain our earlier geometry-projection construction: the parent image is back-projected into 3D using depth, a prescribed 6-DoF camera motion is applied, and the result is rasterized with a z-buffer. These views contain genuine parallax and real occlusion — surfaces that occlude one another under camera movement, and disocclusion holes where previously hidden geometry is revealed.
% I will supplement the second mode in the paper.
% }

\noindent\textbf{Feature puzzles generation.}
Let a parent image $I$ have spatial size $H \times W$. As in Figure~\ref{fig:feature_puzzles_pipeline}B, we sample $N$ overlapping \emph{sub-images} $\{x_n\}_{n=1}^{N}$, each a box in parent coordinates of side length $s \in [s_{\min},s_{\max}]$ (fraction of the shorter side; default $[0.5,0.9]$), resampled to square resolution $R$ so $x_n \in \mathbb{R}^{R \times R \times 3}$. We also write $x_n$ for the box, so $x_i \cap x_j$ is the overlap of two sub-images. A candidate is accepted only if it overlaps an accepted sub-image with IoU $\ge o_{\min}$, so the $N$ sub-images form a connected overlap graph---the condition for cross-view warping.

\noindent\textbf{3D-augmented views.}
Affine crops contain no parallax or occlusion. 
Optionally, following Image-to-Clips~\citep{ma2025puzzles}, we estimate parent depth with MoGe-2~\citep{wang2025moge2}, calibrate each view's camera, and apply 6-DoF augmentation, pulling labels back through depth projection. This optional construction is distinct from the affine crop mapping defined below.

% \david{maybe you can include a clarification below that the exact correspondence claims below refer to the affine crop construction; for 3D-augmented views, correspondence is determined by the estimated depth and projection geometry rather than by the closed-form crop map.}
% \jiahao{Please refer to appendix.}

% \david{Are the reported experiments actually trained with affine crop-and-resize views only, or are MoGe-2 6-DoF augmented views also used?
% If the reported experiments use affine crops only, then I would remove the 3D-augmented paragraph from the main Method and remove or clearly label the Appendix D paragraph as an optional extension not used in the experiments.
% If some experiments really use 6-DoF augmentation, then Appendix A.3 cannot say simply "Our views are restricted to in-plane crop-and-resize"; it needs to specify which branch/experiment uses which mode.
% %
% Appendix A.3 on page 15, where the paper says: "Our views are restricted to in-plane crop-and-resize."
% Appendix D on page 20, where it again discusses: "For 3D-augmented views, rendered RGB and labels use the same estimated depth and virtual-camera projection ..."
% } \jiahao{fix}

\noindent\textbf{The sub-image-to-parent map.}
For the affine crop views, sub-image $x_n$ is related to the parent $I$ by a closed-form affine map $\pi_n$ that we chose when sampling it, so $x_n = I \circ \pi_n$ maps sub-image coordinates to parent coordinates. Unlike a camera pose, $\pi_n$ is never estimated: it is fixed by the sampled box, requires no depth and no camera model, and is invertible in closed form.

\noindent\textbf{Label transfer by pull-back.}
For the affine crop construction, the same map $\pi_n$ that relates a sub-image to the parent RGB transfers labels just as well: rather than annotating each crop independently, we pull back a single parent-space label field $\ell$---which may carry instance masks, semantic classes, or dense teacher features---by composition with $\pi_n$ and resample to crop resolution,
\begin{equation}
    \ell_n = \mathcal{R}_{\mathcal{Y}}\big[\, \ell \circ \pi_n \,\big].
\end{equation}
Here, $\mathcal{R}_{\mathcal{Y}}$ resamples nearest-neighbour for discrete labels (instance identity, semantic class) and bilinearly with re-normalization for dense features. The parent is the single source of truth and each view a restriction of it, so two views cannot disagree: for any point $p$ in the overlap $x_i \cap x_j$,
\begin{equation}
    \ell_i\big(\pi_i^{-1}(p)\big) \;=\; \ell_j\big(\pi_j^{-1}(p)\big) \;=\; \ell(p).
\end{equation}
Here, $\pi_i^{-1}(p)$ is $p$ in sub-image $i$'s local coordinates, so both sides read the same parent label $\ell(p)$ through different crop coordinates---cross-view consistency is exact \emph{by construction} for these shared-parent affine views.

\noindent\textbf{What the parent field is.}
The field $\ell$ takes two forms. \emph{For promptable segmentation}, $\ell$ is the set of SA-1B instance masks $\{\mathcal{A}_m\}$; pulling each mask $\mathcal{A}_m$ back to view $n$ gives that view's binary target $\mathbf{G}_{n,m}$ for instance $m$. Since $m$ indexes the parent field, not a view, it is a \emph{global instance identifier}: channel $m$ is the same object in all $N$ views, so \emph{no} Hungarian matching~\citep{instancegaussian} or cross-view association~\citep{ravi2024sam2} is needed and the prompt--target pairing of Section~\ref{sec:seg_objective} is one-to-one. \emph{For open-vocabulary 3D segmentation}, $\ell$ is a frozen DINOv3 teacher~\citep{simeoni2025dinov3} evaluated \emph{once on the whole parent image} and pulled back by grid sampling. Evaluating before cropping---not per-view---is what guarantees consistency: transformer features change under crop-and-resize, so per-view targets would disagree on the overlap; pull-back instead gives every view a target from the same parent field, so distillation and cross-view consistency share one optimum (Appendix~\ref{app:targets}). Appendix~\ref{app:prompts} gives the augmentation policy, prompt sampling, and annotation filters; Appendix~\ref{app:i2c} contrasts this with the Image-to-Clips augmentation of Puzzles~\citep{ma2025puzzles}, whose synthetic views carry geometric labels from pose fitting and re-rendering. We anneal the minimum crop overlap $o_{\min}$ from $0.7$ to $0.1$, exposing the model first to near-duplicate views and then to progressively larger viewpoint changes with less shared content.

% \paragraph{Overlap curriculum.}
% Training begins with highly overlapping, near-duplicate crops: the required overlap $o_{\min}$ starts at $0.7$ and is linearly annealed to $0.1$. The model is therefore first exposed to correspondence under small transformations and gradually encounters views with less shared content.

% \david{I'm confused here. please confirm which view-generation mode is actually used for each reported experiment/task. If the 3D-augmented views contribute to the reported results, the method and appendix need to describe that protocol explicitly and the "exact correspondence" claim must remain restricted to the affine component. If the reported results use only affine crops, the 3D augmentation should be clearly identified as optional and not part of the core experimental configuration.}

\subsection{Pretrained   Multi-View Backbone}
\label{sec:backbone}

% \david{'Plug-and-play' and especially 'arbitrary feed-forward reconstruction models' are not demonstrated with a single VGGT-Omega instantiation. Either add a second backbone experiment or narrow the claim.} \jiahao{I will add the LingBot-Map results here to demonstrate the 'Plug-and-play'.}

\begin{figure}[t]
    \centering
    \includegraphics[width=0.98\linewidth]{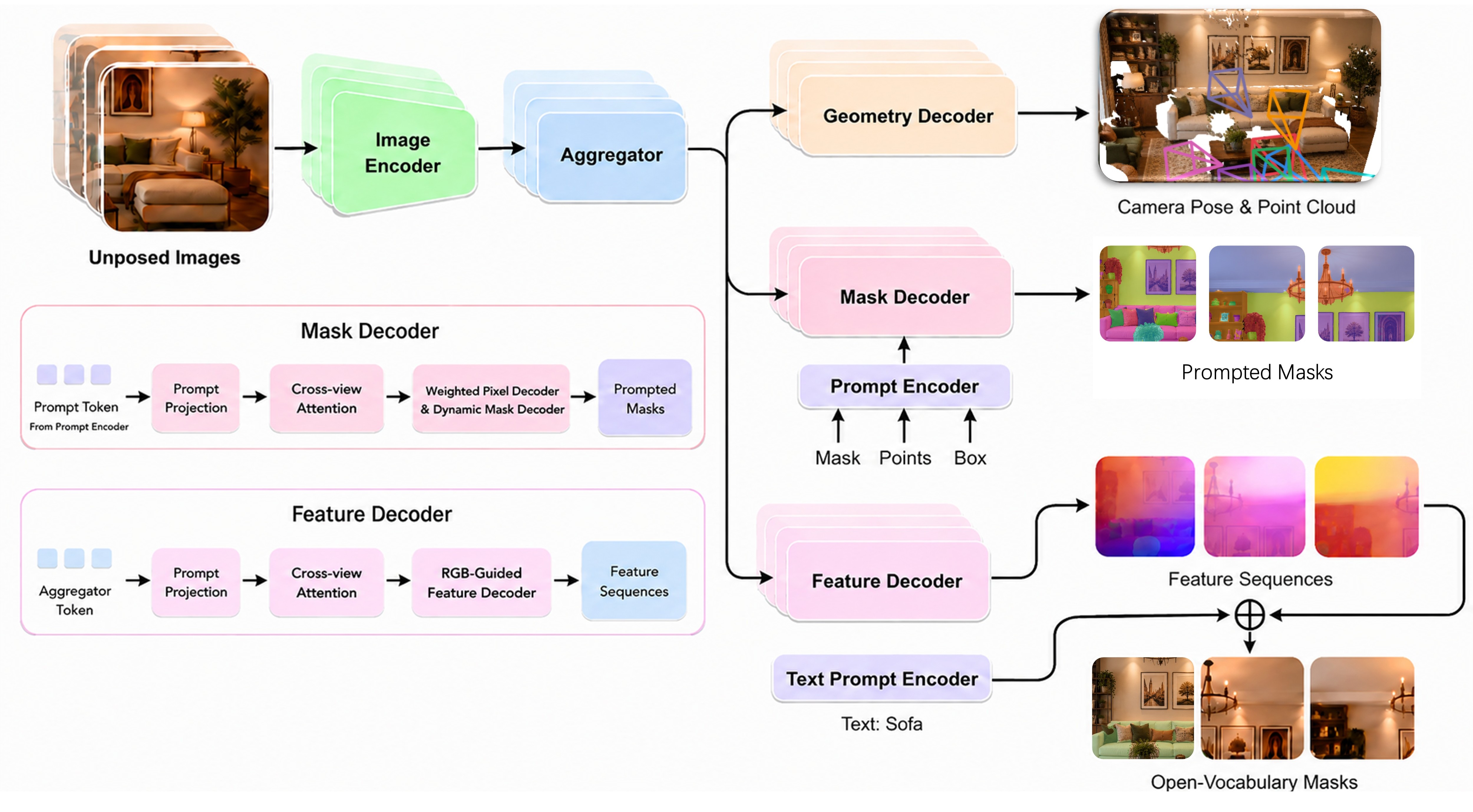}
    \vspace{-0.4cm}
    \caption{\textbf{Feature Puzzles network architecture.}
    A shared \emph{Image Encoder} + \emph{Aggregator} produce tokens feeding three decoders: \emph{Geometry} (poses, 3D points), prompt-conditioned \emph{Mask} (mask/point/box $\to$ masks), and \emph{Feature} ($\oplus$ text-prompt $\to$ open-vocab masks). Mask and feature decoders share cross-view attention and a weighted pixel decoder. 
    % \jiahao{add icon for training and freeze module.}
    }
    \label{fig:network_architecture}
    \vspace{-0.2cm}
\end{figure}

As shown in Figure~\ref{fig:network_architecture}, we build on a pretrained VGGT-Omega~\citep{vggtomega} multi-view transformer.
We keep its early blocks frozen and adapt late blocks with LoRA~\citep{hu2022lora}, preserving the original depth and camera heads at inference.
The segmentation heads consume a shared representation $\mathbf{F}_n$ obtained by fusing cached multi-layer tokens.
Adapters are initialized to preserve the pretrained backbone at the start of training, enabling two-pass inference with adapters off for reconstruction and on for segmentation.
We further evaluate the same adaptation recipe on LingBot-Map~\citep{lingbotmap}, a streaming reconstruction backbone with a different multi-view attention design, providing evidence that the approach is not specific to VGGT-Omega.
Layer choice, fusion, and initialization are detailed in Appendix~\ref{app:backbone}.

\subsection{Promptable Multi-View Segmentation}
\label{sec:promptable}

\noindent\textbf{Architecture.}
As shown in Figure~\ref{fig:network_architecture} (\emph{Mask Decoder}), given the sub-images $\{x_n\}$ and a set of prompts (masks, boxes, or labelled points), the model predicts masks for $M$ instances across all $N$ views in a single forward pass. 
A SAM-style prompt encoder~\citep{sam} encodes each prompt into a dense coverage map $\mathbf{C}_{n,m}$ and a sparse prompt token; during training prompts may be supplied in one or multiple views. The coverage map injects the prompt into the shared representation $\mathbf{F}_n$ from the backbone through a feature-wise linear projection module,
\begin{equation}
    \mathbf{F}'_{n,m}
    = \gamma(\mathbf{C}_{n,m}) \odot \mathbf{F}_n
    + \beta(\mathbf{C}_{n,m}),
\end{equation}
where $\gamma$ and $\beta$ are learnable projections producing a per-channel scale and shift, yielding an instance-conditioned feature $\mathbf{F}'_{n,m}$ (unprompted view--instance pairs pass through unchanged). The instance-conditioned tokens then pass through cross-view self-attention---which allows prompt-conditioned information from one view to propagate to the others---and SAM-style two-way attention between image and object tokens, denoted $\mathbf{H}_{n,m}=\mathrm{Attn}(\mathbf{F}'_{n,m})$ with its image and object outputs $\mathbf{H}_{n,m}^{\mathrm{img}},\mathbf{H}_{n,m}^{\mathrm{obj}}$. 
A weighted pixel decoder and a per-instance kernel head then produce the mask logits,
\begin{equation}
\mathbf{D}_n = \mathrm{PixelDecoder}(\mathbf{H}_{n,m}^{\mathrm{img}}), \quad
\mathbf{k}_{n,m} = \mathrm{KernelHead}(\mathbf{H}_{n,m}^{\mathrm{obj}}), \quad
\mathbf{Y}_{n,m} = \big\langle \mathbf{D}_n,\, \mathbf{k}_{n,m} \big\rangle.
\end{equation}
Auxiliary heads predict mask IoU. Appendix~\ref{app:promptable} details the token layout and attention blocks.

\noindent\textbf{Training objective.}
\label{sec:seg_objective}
Every prompt corresponds one-to-one to a ground-truth mask, and targets are warped into every crop---including crops where the prompt was not given---so the objective is a plain weighted sum,
\begin{equation}
    \mathcal{L}_{\mathrm{seg}}
    = \lambda_{\mathrm{bce}}\mathcal{L}_{\mathrm{bce}}
    + \lambda_{\mathrm{dice}}\mathcal{L}_{\mathrm{dice}}
    + \lambda_{\mathrm{iou}}\mathcal{L}_{\mathrm{iou}}
    + \lambda_{\mathrm{cons}}\mathcal{L}_{\mathrm{cons}}
    + \lambda_{\mathrm{pt}}\mathcal{L}_{\mathrm{pt}},
\end{equation}
combining: 
$\mathcal{L}_{\mathrm{bce}}$ and $\mathcal{L}_{\mathrm{dice}}$ for pixel-wise mask supervision (binary cross-entropy and Dice), 
$\mathcal{L}_{\mathrm{iou}}$ for IoU regression on a per-view confidence head,
$\mathcal{L}_{\mathrm{cons}}$ as an auxiliary cross-view token regularizer that pulls object tokens of the same instance toward their cross-view centroid, and 
$\mathcal{L}_{\mathrm{pt}}$ for point-prompt supervision. Per-term details are in Appendix~\ref{app:promptable}.

% \david{maybe a reviewer could see the limitation of supporting the 'multi-view prompting capabilitiy. NVOS and SPIn-NeRF use a single reference-view mask prompt, while the ablation only shows that removing multi-view prompting during training reduces performance. If “prompts from multiple views jointly improve or propagate segmentation” is intended as an empirical contribution rather than only an architectural capability, we should have a direct one-prompt versus multi-prompt result.}
% \jiahao{add the ablation study to the appendix.}

% \david{I'm not sure if you already addressed this. The current promptable ablations still do not directly isolate the contribution of Feature Puzzles itself. You test LoRA, cross-view attention, overlap difficulty, multi-view prompting, but there is no promptable equivalent of a single-view/no-puzzle training baseline. a reviewer could still ask whether the shared-parent puzzle supervision is necessary for this branch rather than the pretrained backbone plus cross-view decoder being sufficient.}
% \jiahao{add the ablation study to the appendix.}

\subsection{Open-Vocabulary 3D Segmentation}
\label{sec:openvocab}

% \david{if feasible Separate the contribution into (i) single-image DINO feature pretraining and (ii) in-domain semantic fine-tuning. The strongest Table 3 numbers use semantic supervision for ScanNet/ScanNet200/ScanNet++ vocabularies, so those results should not be used as the primary evidence for zero-shot open-vocabulary generalisation.}
% \jiahao{Agreed, and the problem is slightly worse than stated. Ours (SN200, SN++)‡ is fine-tuned on the ScanNet200 / ScanNet++ vocabularies training data and then evaluated on ScanNet 10/15/19, ScanNet200 and ScanNet++ — on those subsets it is not open-vocabulary at all, it is closed-set supervised. It is currently bolded as the best result and sits amid zero-shot methods.
% 1.Split the table into zero-shot / open-vocabulary and in-domain supervised blocks. The † row becomes the primary result; no bolding or comparison across blocks.
% 2.Reorder the narrative. §4.4 currently opens with the ‡ row; it will open with †, and ‡ will be demoted to "the same representation is also a strong initialization for in-domain fine-tuning."
% 3.Drop "(fair)" from the caption. †~SA-1B zero-shot evaluation (fair) — asserting fairness in a caption reads as defensive; we will describe the protocol objectively instead.}

\noindent\textbf{RGB-guided dense feature head.}
\label{sec:feature_head}
The Feature Decoder in Figure~\ref{fig:network_architecture} takes the shared representation $\mathbf{F}_n$ from the backbone, mixes the $N$ views with cross-view attention, and upsamples with RGB guidance at each stage to restore image-aligned boundaries, yielding a dense feature map $\widehat{\mathbf{F}}_n$ aligned with the frozen DINOv3~\citep{simeoni2025dinov3}/Talk2DINO~\citep{barsellotti2025talk2dino} space.

\noindent\textbf{Stage 1: single-image pre-training.}
We first train the head on SA-1B puzzle crops with
\begin{equation}
    \mathcal{L}_{\mathrm{stage1}}
    = \lambda_{\mathrm{feat}}\mathcal{L}_{\mathrm{feat}}
    + \lambda_{\mathrm{warp}}\mathcal{L}_{\mathrm{warp}}
    + \lambda_{\mathrm{gram}}\mathcal{L}_{\mathrm{gram}},
\end{equation}
where $\mathcal{L}_{\mathrm{feat}}$ is the mean pixel-wise cosine distance between the student $\widehat{\mathbf{F}}_n$ and the pulled-back DINOv3 teacher of that view, and the other two terms exploit the exact crop geometry of the shared-parent affine views. The cross-view warping term maps a student feature from view $i$ onto view $j$ over their overlap and pulls it toward a stop-gradient target:
\begin{equation}
    \mathcal{L}_{\mathrm{warp}}
    = \frac{1}{|\mathcal{P}|}\sum_{(i,j)\in\mathcal{P}}
      \frac{1}{|\Omega_{ij}|}\sum_{q \in \Omega_{ij}}
      \Big(1 - \cos\big(\widehat{\mathbf{F}}_i\big(\pi_i^{-1}\!\circ\pi_j(q)\big),\;
      \mathrm{sg}\big[\widehat{\mathbf{F}}_j(q)\big]\big)\Big),
\end{equation}
where $\mathcal{P}$ is the set of overlapping view pairs, $\Omega_{ij}$ their overlap, $q$ a pixel in view $j$, and $\pi_i^{-1}\!\circ\pi_j$ maps $q$ from view $j$ into view $i$ so both cosine arguments sit at the same parent location; $\mathrm{sg}[\cdot]$ blocks the gradient through the target. The Gram anchoring term matches pairwise similarities. We sample $K$ pixels $\{p_k\}$ from a view and stack the L2-normalized student features $\widehat{\mathbf{F}}_n(p_k)$ and the pulled-back teacher features $\mathbf{F}_t(p_k)$ at the same pixels as the rows of $\widehat{\mathbf{S}}=[\widehat{\mathbf{F}}_n(p_1);\ldots;\widehat{\mathbf{F}}_n(p_K)]$ and $\mathbf{S}_{T}=[\mathbf{F}_t(p_1);\ldots;\mathbf{F}_t(p_K)]$, then match the $K\!\times\!K$ Gram matrices
\begin{equation}
    \mathcal{L}_{\mathrm{gram}}
    = \left\|
    \widehat{\mathbf{S}}\widehat{\mathbf{S}}^\top
    - \mathbf{S}_{T}\mathbf{S}_{T}^\top
    \right\|_F^2 ,
\end{equation}
%preserving sharp cross-boundary dissimilarity and tying different scales to a common similarity geometry. 
which is designed to preserve cross-boundary feature dissimilarity and tie different scales to a common similarity geometry.
Weights and sampling details are in Appendix~\ref{app:openvocab}.

\noindent\textbf{Stage 2: in-domain text-aligned fine-tuning.}
\label{sec:stage2}
Stage~1 aligns $\widehat{\mathbf{F}}_n$ with DINOv3 but not with language. We therefore add per-pixel text-aligned cross-entropy using labels warped from the parent image. For pixel $p$ and class $k$,
$z_{p,k}=\cos(\widehat{\mathbf{f}}_p,\mathbf{e}_k)/\tau$ with $\tau=0.07$, where $\mathbf{e}_k$ is obtained from a frozen CLIP text tower and mapped into DINOv3 space by frozen Talk2DINO. Stage-1 distillation is retained to preserve DINOv3 alignment, while the new loss introduces explicit supervision for the in-domain semantic classes.

\noindent\textbf{3D lifting and classification.}
At inference, each RGB frame produces a dense feature map. Depth and camera pose are used to unproject feature pixels into 3D, where features are averaged within a $0.02$\,m voxel grid to obtain one feature $\mathbf{f}_{\mathrm{voxel}}$ per voxel. Each query point takes its nearest voxel feature and is classified by
$\hat{y}=\arg\max_k\cos(\mathbf{f}_{\mathrm{voxel}},\mathbf{e}_k)$, where $\mathbf{e}_k$ is obtained from the frozen CLIP/Talk2DINO text mapping. The entire process is feed-forward and requires no scene-specific training or test-time optimization.
% \david{The current benchmark uses ground-truth geometry, so the results measure the quality of the learned semantic feature field given geometry, rather than end-to-end RGB-only 3D reconstruction and semantic understanding. This does not necessarily need more detail in this Method subsection, but it must be stated explicitly in the corresponding experimental protocol before making comparisons with end-to-end RGB/RGB-D systems.}
% \jiahao{Fix.}

% \subsection{Implementation Details}
% \david{Move completely to appendix}
% Both tasks share the cached VGGT-Omega blocks $\{4,11,17,23\}$, a working dimension of 768, bfloat16 mixed precision, AdamW with cosine decay and linear warm-up, and training on eight A800 GPUs. Promptable segmentation uses six views; open-vocabulary pre-training uses six $256\times256$ crops and predicts $128\times128$ features. Appendix~\ref{app:impl} lists the remaining hyper-parameters.

%--------------------------------------------------------------------------------------
%--------------------------------------------------------------------------------------
%--------------------------------------------------------------------------------------
\section{Experiments}
Feature Puzzles extends feed-forward reconstruction with scene understanding. We evaluate promptable segmentation on NVOS and SPIn-NeRF (Section~\ref{sec:exp_nvos}), instance tracking on ScanNet++ (Section~\ref{sec:exp_scannet}), OOD tracking on nuScenes-mini (Section~\ref{sec:exp_nuscenes}), and open-vocabulary 3D segmentation following EmbodiedSplat (Section~\ref{sec:open_vocabulary_segmentation}).

\subsection{Promptable Multi-View Segmentation}
\label{sec:exp_nvos}
We evaluate promptable multi-view segmentation on NVOS~\citep{ren2022nvos} and SPIn-NeRF~\citep{mirzaei2023spinnerf} following the SAGA protocol, using a single reference-view mask prompt. Task-specific training uses only single-image SA-1B-style masks, with no video or real multi-view supervision. As shown in Table~\ref{tab:nvos_spinerf}, our feed-forward model achieves the best mIoU on NVOS ($93.2\%$) and outperforms SAM2-Video on SPIn-NeRF ($88.5\%$ vs.~$86.6\%$), while remaining below MV-SAM and the strongest per-scene methods there. Across both datasets, inference requires no scene-specific optimization and generalizes directly to unseen scenes.

% We evaluate promptable multi-view segmentation on NVOS~\citep{ren2022nvos} and SPIn-NeRF~\citep{mirzaei2023spinnerf} following the SAGA evaluation protocol. 
% Our method uses a single reference-view mask prompt (the simplest SAGA variant) and 
% its task-specific segmentation training uses only single-image SA-1B-style masks, with no task-specific video or multi-view supervision--unlike SAM2-Video (trained on video) and the per-scene optimization baselines (which fit each test scene individually). As shown in Table~\ref{tab:nvos_spinerf}, on NVOS we achieve the best mIoU ($93.2\%$) across all methods---per-scene and generalization alike---and mAcc $98.3\%$ that beats the generalization baselines MV-SAM ($97.5\%$) and SAM2-Video ($94.6\%$), though per-scene methods reach slightly higher mAcc ($98.4$--$98.6\%$). On SPIn-NeRF (3D-Scene-Segmentation-HQ, 5 scenes, single reference-view prompt) we report $88.5\%$ mIoU and $97.7\%$ mAcc: below per-scene methods ($90.7$--$94.5\%$) and MV-SAM ($92.9\%$), but above SAM2-Video ($86.6\%$) and with the best mAcc among generalization methods.
% Thus, on NVOS our feed-forward model achieves higher mIoU than the compared per-scene optimization methods, while on SPIn-NeRF it remains below the strongest per-scene and generalization baselines in mIoU. Across both datasets, inference requires no scene-specific optimization and generalizes directly to unseen scenes from task-specific training based only on single-image SA-1B masks.

\begin{table}[t]
\centering
\small
\setlength{\tabcolsep}{3pt}
\caption{\textbf{Promptable multi-view segmentation on NVOS and SPIn-NeRF.} Per-scene methods fit each test scene; generalization methods use one forward pass. 
Ours uses only SA-1B-style masks for task-specific segmentation training.}
\label{tab:nvos_spinerf}
\vspace{-0.2cm}
\resizebox{\columnwidth}{!}{%
\begin{tabular}{lllcccc}
\toprule
Category & Method & Train & \multicolumn{2}{c}{NVOS} & \multicolumn{2}{c}{SPIn} \\
\cmidrule(lr){4-5} \cmidrule(lr){6-7}
 & & & mIoU (\%) & mAcc (\%) & mIoU (\%) & mAcc (\%) \\
\midrule
\multirow{5}{*}{\makecell[l]{Per-scene\\opt.}} & SPIn-NeRF~\citep{mirzaei2023spinnerf} & NeRF & -- & -- & 90.7 & 98.8 \\
 & SA3D~\citep{cen2023sa3d} & NeRF+SAM & 91.1 & 98.4 & 92.4 & 98.8 \\
 & SAGA~\citep{cen2023saga} & 3DGS+SAM & 92.6 & 98.6 & 93.7 & 99.2 \\
 & SA3D-GS~\citep{cen2025sa3dgs} & 3DGS+SAM & 92.7 & 98.5 & 93.4 & 99.1 \\
 & OmniSeg3D~\citep{ying2024omniseg3d} & 3DGS+SAM & 92.8 & 98.6 & 94.5 & 99.3 \\
\midrule
\multirow{3}{*}{General.} & SAM2~\citep{ravi2024sam2} &  SA-V (video) & 88.7 & 94.6 & 86.6 & 93.6 \\
 & MV-SAM~\citep{jeong2026mvsam} & SA-1B (single image) & 92.1 & 97.5 & \textbf{92.9} & 97.1 \\
 & \textbf{Ours} & SA-1B (single image) & \textbf{93.2} & \textbf{98.3} & 88.5 & \textbf{97.7} \\
\bottomrule
\end{tabular}%
}
\vspace{-0.2cm}
\end{table}

\subsection{Instance Spatial Tracking}
\label{sec:exp_tracking}
\emph{ScanNet++ evaluation.}
\label{sec:exp_scannet}
We follow the IGGT~\citep{li2026iggt} protocol on ScanNet++~\citep{yeshwanth2023scannet++}, propagating a reference mask to all frames in one forward pass over 19 instances from 7 scenes. IGGT is trained on ScanNet++ RGB-D and is therefore in-distribution, whereas our task-specific segmentation training uses only SA-1B and is OOD to ScanNet++. As shown in Table~\ref{tab:tracking_merged}, our method achieves $63.60\%$ mIoU and $87.50\%$ T-SR, approaching IGGT ($73.02\%$ / $98.90\%$) while outperforming SAM2* and SpaTracker+SAM.

\emph{OOD evaluation on nuScenes-mini.}
\label{sec:exp_nuscenes}
We adapt the same protocol to nuScenes-mini~\citep{caesar2020nuscenes} (9 scenes, 6 surround cameras), using projected 3D boxes as pseudo-masks. Here, all methods are OOD: IGGT is trained on ScanNet++ RGB-D, while our task-specific segmentation training still uses only SA-1B. 
Despite wide baselines and minimal frustum overlaps across the six cameras, our method in one forward pass achieves $4.78\%$ T-mIoU$_\mathrm{all}$, $20.97\%$ T-mIoU$_\mathrm{vis}$, and the highest T-SR$_\mathrm{vis}$ of $88.89\%$. IGGT obtains higher T-SR$_\mathrm{all}$ through HDBSCAN matching but lower mIoU. See Appendix~\ref{app:vis_promptable_seg} for detailed visual comparisons on ScanNet++ and nuScenes-mini.

\begin{table*}[t]
\centering
\small
\setlength{\tabcolsep}{4pt}
\caption{\textbf{Instance spatial tracking on ScanNet++ and nuScenes-mini.} IGGT is in-distribution on ScanNet++ but OOD on nuScenes. Ours is OOD with respect to task-specific training on both benchmarks and uses only SA-1B for task-specific segmentation training. T-SR denotes temporal success rate; nuScenes reports all-frame (all) and visible-frame (vis) metrics.
}
\label{tab:tracking_merged}
\vspace{-0.2cm}
\resizebox{\textwidth}{!}{%
\begin{tabular}{llcc cccc}
\toprule
\multirow{2}{*}{Method} & \multirow{2}{*}{Training Data} & \multicolumn{2}{c}{ScanNet++} & \multicolumn{4}{c}{nuScenes-mini} \\
\cmidrule(lr){3-4} \cmidrule(lr){5-8}
 & & mIoU (\%) & T-SR (\%) & T-mIoU$_\mathrm{all}$ & T-mIoU$_\mathrm{vis}$ & T-SR$_\mathrm{all}$ & T-SR$_\mathrm{vis}$ \\
\midrule
SpaTracker+SAM~\citep{li2026iggt} & Kubric synth. (video) & 16.15 & 23.68 & -- & -- & -- & -- \\
CoTracker3+SAM~\citep{karaev2025cotracker3} & Kubric synth. (video) & -- & -- & 4.99 & 18.15 & 0.00 & 17.41 \\
SAM2*~\citep{ravi2024sam2} & SA-V (video) & 44.16 & 57.89 & 4.48 & 20.71 & 0.00 & 28.70 \\
IGGT~\citep{li2026iggt} & ScanNet++ RGB-D & \textbf{73.02} & \textbf{98.90} & 2.34 & 9.46 & 59.26 & 78.89 \\
\midrule
\textbf{Ours} & SA-1B (single image) & 63.60 & 87.50 & \textbf{4.78} & \textbf{20.97} & \textbf{47.13} & \textbf{88.89} \\
\bottomrule
\end{tabular}%
 }
% \vspace{-0.1cm}
\end{table*}

\subsection{Open-Vocabulary 3D Segmentation}
\label{sec:open_vocabulary_segmentation}
We follow the EmbodiedSplat protocol~\citep{embodiedsplat2026} on ScanNet, ScanNet200, and ScanNet++, using 11 ScanNet and 4 ScanNet++ scenes with stride~10. Features are voxel-averaged at $0.02$\,m, lifted with ground-truth geometry, and classified using Talk2DINO text embeddings. Thus, the benchmark measures semantic feature quality under fixed geometry. We compare with per-scene 3DGS methods~\citep{qin2024langsplat,shi2024legaussians,wu2024opengaussian,occamslgs2025,drsplat2025,instancegaussian2025,onlinelangsplat2024} and feed-forward EmbodiedSplat. See Appendix~\ref{app:vis_promptable_seg} for coherent semantic reconstruction across diverse large-scale environments.

\emph{Zero-shot.}
As shown in Table~\ref{tab:embodiedsplat_acc}, our Stage-1 model uses only SA-1B task-specific training with no in-domain labels, outperforming the listed per-scene 3DGS methods on ScanNet-15/19 while remaining below EmbodiedSplat trained on in-domain RGB-D scans.
% As shown in Table~\ref{tab:embodiedsplat_acc}, our Stage-1 model uses only SA-1B for task-specific training ($^\dagger$), with no in-domain labels, and achieves higher mIoU than the listed per-scene 3DGS methods on ScanNet-15 and ScanNet-19. It remains below EmbodiedSplat, which is trained on in-domain RGB-D scans.

\emph{In-domain fine-tuning.}
Table~\ref{tab:embodiedsplat_acc} also reports results after fine-tuning the same representation on ScanNet200/ScanNet++ vocabularies ($^\ddagger$). Because these classes are explicitly supervised during fine-tuning, this setting is closed-set rather than zero-shot open-vocabulary. In this setting, it exceeds EmbodiedSplat on mIoU on all five subsets (e.g., SN200 $47.79$ \emph{vs.} $34.75$, SN++ $54.68$ \emph{vs.} $44.03$), showing that the SA-1B-pretrained features provide an effective initialization for in-domain training.

\emph{Efficiency.}
For the speed benchmark, we measure the model's full feed-forward pipeline, including its own reconstruction and semantic prediction, without using ground-truth depth. On a 363-image scene, inference takes $\sim25.9$\,s. This is substantially faster than the reported scene-processing times of per-scene 3DGS methods ($\sim2$--$6$\,hr) and about $18\times$ faster than EmbodiedSplat ($8$\,min).

% \david{Please check the changes above. the zero-shot and in-domain rows must remain clearly separated in interpretation. The SA-1B model is the genuine zero-shot/open-vocabulary result and remains below EmbodiedSplat. The fine-tuned row exceeds EmbodiedSplat in mIoU, but because it uses the evaluated semantic classes during training, that comparison is closed-set.} \jiahao{I refine the table now.}

% \david{I'm not sure about the comparison with EmbodiedSplat. The 25.9 result is obtained in an evaluation that lifts features using ground-truth geometry, whereas EmbodiedSplat performs semantic 3D scene construction. Therefore, "18X faster than EmbodiedSplat" is not a clean end-to-end comparison unless the timing components are matched. I would claim this as substantially faster semantic feature inference under the stated evaluation protocol.}
% \jiahao{Ground-truth geometry is used for semantic evaluation, whereas the inference speed test measures the network's own feed-forward inference time, following the same setting as the generalizable inference version of EmbodiedSplat.}

\begin{table}[t]
\centering
\caption{\textbf{3D semantic segmentation on the EmbodiedSplat benchmark.}
$^\dagger$~zero-shot evaluation with SA-1B task-specific training only;
$^\ddagger$~closed-set evaluation after in-domain fine-tuning on the evaluated semantic classes.
``cls'' denotes class. Bold = best, underline = second best.}
\label{tab:embodiedsplat_acc}
\vspace{-0.2cm}
\resizebox{\textwidth}{!}{%
\begin{tabular}{lccccccccccccc}
\toprule
\multirow{2}{*}{Method} & \multicolumn{6}{c}{ScanNet} & \multicolumn{2}{c}{ScanNet200} & \multicolumn{2}{c}{ScanNet++} & \multirow{2}{*}{\makecell{Recon.\\time}} & \multirow{2}{*}{\makecell{Per-Scene/\\Gen.}} & \multirow{2}{*}{\makecell{on/\\off}} \\
\cmidrule(lr){2-7}\cmidrule(lr){8-9}\cmidrule(lr){10-11}
 & \multicolumn{2}{c}{10 cls} & \multicolumn{2}{c}{15 cls} & \multicolumn{2}{c}{19 cls} & \multicolumn{2}{c}{70 cls} & \multicolumn{2}{c}{20 cls} & & & \\
\cmidrule(lr){2-3}\cmidrule(lr){4-5}\cmidrule(lr){6-7}\cmidrule(lr){8-9}\cmidrule(lr){10-11}
 & mIoU & mAcc & mIoU & mAcc & mIoU & mAcc & mIoU & mAcc & mIoU & mAcc & & & \\
\midrule

\multicolumn{14}{l}{\textit{In-domain / benchmark-specific methods}} \\
\midrule
LangSplat~\citep{qin2024langsplat} & 6.52 & 20.11 & 3.25 & 13.16 & 1.34 & 7.44 & 0.72 & 4.39 & 2.21 & 10.18 & $\sim$6 hr & \multirow{3}{*}{Per-Scene} & \multirow{2}{*}{Offline} \\
LEGaussians~\citep{shi2024legaussians} & 6.79 & 18.13 & 4.13 & 15.49 & 2.53 & 5.67 & 1.39 & 5.45 & 2.93 & 9.34 & $\sim$6 hr & & \\
Online-LangSplat~\citep{onlinelangsplat2024} & 7.13 & 21.56 & 3.89 & 14.52 & 3.45 & 8.97 & 2.45 & 4.12 & 4.51 & 11.34 & 5.4 min & & Online \\
\midrule
OpenGaussian~\citep{wu2024opengaussian} & 29.50 & 44.61 & 23.74 & 39.14 & 22.52 & 35.02 & 15.15 & 25.66 & 25.65 & 37.03 & $\sim$2.5 hr & \multirow{4}{*}{Per-Scene} & \multirow{4}{*}{Offline} \\
Occam's LGS~\citep{occamslgs2025} & 42.14 & 70.28 & 35.04 & 63.71 & 30.49 & 57.91 & 20.32 & 40.49 & 34.08 & 61.19 & $\sim$2 hr & & \\
Dr.~Splat~\citep{drsplat2025} & 39.21 & 66.66 & 31.84 & 60.58 & 28.38 & 55.85 & 19.29 & 33.84 & 39.85 & 58.34 & $\sim$2 hr & & \\
InstanceGaussian~\citep{instancegaussian2025} & 29.77 & 52.32 & 28.79 & 50.07 & 26.57 & 48.63 & 23.20 & 38.32 & 29.98 & 47.47 & $\sim$3 hr & & \\
\midrule
EmbodiedSplat~\citep{embodiedsplat2026} & \underline{57.41} & \underline{82.45} & \underline{55.18} & \textbf{80.27} & \underline{52.12} & \underline{75.66} & \underline{34.75} & 52.36 & 44.03 & 66.27 & 8 min & \multirow{2}{*}{Generalizable} & \multirow{2}{*}{Online} \\
EmbodiedSplat-fast~\citep{embodiedsplat2026} & 51.05 & 80.15 & 46.92 & 77.15 & 43.89 & 72.73 & 32.43 & \underline{58.14} & \underline{51.09} & \textbf{78.68} & 1 min 10 s & & \\
\midrule
\textbf{Ours (SN200, SN++)}$^\ddagger$ & \textbf{68.03} & \textbf{83.32} & \textbf{64.38} & \underline{80.11} & \textbf{60.95} & \textbf{78.90} & \textbf{47.79} & \textbf{67.74} & \textbf{54.68} & 66.97 & $\sim$25.9 s & Generalizable & Online \\

\midrule
\multicolumn{14}{l}{\textit{Zero-shot / Open-vocabulary}} \\
\midrule
\textbf{Ours (SA-1B)}$^\dagger$ & 41.09 & 58.55 & 35.87 & 50.73 & 30.50 & 45.93 & 13.79 & 23.79 & 30.12 & 49.23 & $\sim$25.9 s & Generalizable & Online \\
\bottomrule
\end{tabular}%
}
%\vspace{-0.2cm}
\end{table}

\begin{figure}[t]
\centering
\includegraphics[width=\linewidth]{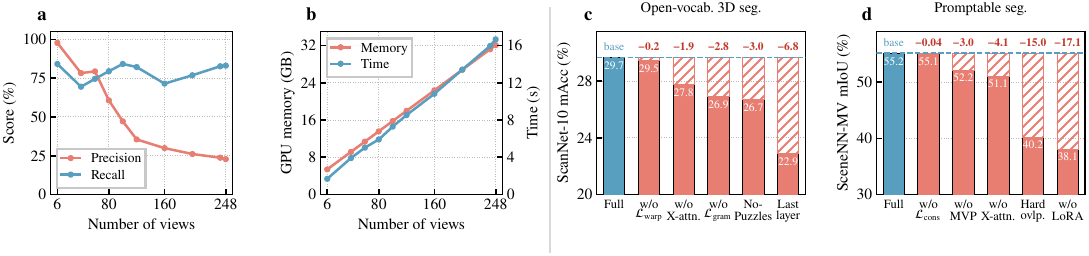}
\vspace{-0.3cm}
% \caption{\textbf{Ablation study.}
% \textbf{(a,\,b)}~Precision/recall, memory, and runtime versus number of views.
% The long-view experiment is a stress test: memory and runtime scale approximately linearly, while precision degrades substantially at large view counts.
% \textbf{(c,\,d)}~Component ablations for open-vocabulary (ScanNet-10 mAcc) and promptable segmentation (SceneNN-MV mIoU); numbers show the drop from the full model.}
\caption{\textbf{Ablation study.}
\textbf{(a,\,b)}~Precision/recall, memory, and runtime versus view count.
The long-view stress test shows near-linear memory/runtime scaling, while precision degrades at large view counts.
\textbf{(c,\,d)}~Component ablations for open-vocabulary (ScanNet-10 mAcc) and promptable segmentation (SceneNN-MV mIoU); numbers indicate the drop from the full model.}
\label{fig:ablation}
%\vspace{-0.2cm}
\end{figure}

\begin{table}[t]
\centering
\caption{\textbf{Multi-backbone generalization.} The same head and task-specific SA-1B training are applied to an offline and a streaming reconstruction backbone, without in-domain fine-tuning.}
\label{tab:backbone_ablation}
% \vspace{-0.2cm}
\resizebox{\textwidth}{!}{%
\begin{tabular}{ll cccccccc cccc}
\toprule
\multirow{4}{*}{Backbone} & \multirow{4}{*}{Multi-view architecture} & \multicolumn{8}{c}{Open-vocabulary 3D segmentation} & \multicolumn{4}{c}{Instance spatial tracking} \\
\cmidrule(lr){3-10}\cmidrule(lr){11-14}
 & & \multicolumn{6}{c}{ScanNet} & \multicolumn{2}{c}{ScanNet200} & \multicolumn{4}{c}{nuScenes-mini} \\
\cmidrule(lr){3-8}\cmidrule(lr){9-10}\cmidrule(lr){11-14}
 & & \multicolumn{2}{c}{10 cls} & \multicolumn{2}{c}{15 cls} & \multicolumn{2}{c}{19 cls} & \multicolumn{2}{c}{70 cls} & \multirow{2}{*}{T-mIoU$_\mathrm{all}$} & \multirow{2}{*}{T-mIoU$_\mathrm{vis}$} & \multirow{2}{*}{T-SR$_\mathrm{all}$} & \multirow{2}{*}{T-SR$_\mathrm{vis}$} \\
\cmidrule(lr){3-4}\cmidrule(lr){5-6}\cmidrule(lr){7-8}\cmidrule(lr){9-10}
 & & mIoU & mAcc & mIoU & mAcc & mIoU & mAcc & mIoU & mAcc & & & & \\
\midrule
LingBot-Map & Online, streaming & 27.38 & 40.26 & 19.45 & 29.88 & 15.69 & 27.21 & 5.67 & 11.38 & 3.68 & 13.21 & 41.98 & 78.90 \\
\addlinespace
VGGT-Omega~(default) & Offline, all-view attention & \textbf{41.09} & \textbf{58.55} & \textbf{35.87} & \textbf{50.73} & \textbf{30.50} & \textbf{45.93} & \textbf{13.79} & \textbf{23.79} & \textbf{4.78} & \textbf{20.97} & \textbf{47.13} & \textbf{88.89} \\
\bottomrule
\end{tabular}%
}
%\vspace{-0.3cm}
\end{table}

%--------------------------------------------------------------------------------------
%--------------------------------------------------------------------------------------
%--------------------------------------------------------------------------------------
\section{Ablation Study}
% \noindent\textbf{Speed and memory.}
% We study promptable-model scalability with view count (Figure~\ref{fig:ablation}a--b), using a 6-view operating point and a long-view stress test on SceneNN-MV scene~005 with up to 248 frames. At 6 views, the model achieves $97.8\%$ precision and $84.2\%$ recall. Memory and runtime scale approximately linearly, from $5.40$\,GB / $1.68$\,s at 6 views to $32.11$\,GB / $16.68$\,s at 248 views. However, semantic scalability is weaker: precision drops from $78.3\%$ at 40 views to $22.8\%$ at 248 views, while recall remains relatively stable ($69.5$--$83.1\%$). Thus, the model scales computationally to long sequences, but perceptual scalability remains a limitation.
\paragraph{Speed and memory.} We study how a single forward pass of the promptable model scales with the number of input views (Figure~\ref{fig:ablation}a--b). We report a 6-view operating point from the view-count sweep and a long-view stress test on SceneNN-MV scene~005 (40--248 frames), whose longest contiguous sequence contains 248 frames.
At 6 views the model achieves $97.8\%$ precision and $84.2\%$ recall. On the long sequence, precision falls steadily as absent-view false positives accumulate, from $78.3\%$ at 40 views to $22.8\%$ at 248 views, while recall stays comparatively stable ($69.5$--$83.1\%$). 
GPU memory and runtime both grow approximately linearly with the view count, from $5.40$\,GB / $1.68$\,s at 6 views to $32.11$\,GB / $16.68$\,s at 248 views.
Although all 248 frames fit in a single forward pass, the substantial precision degradation at large view counts indicates that perceptual scalability, rather than memory alone, remains a limitation.

% \david{clarification about my edit. the long-view result should be presented as both scalability evidence and a limitation. Processing 248 frames in one pass demonstrates computational scalability, but precision falls to 22.8\%. The paper should therefore avoid implying that successful processing of 248 views means semantic performance scales equally well.}

\noindent\textbf{Component ablations.}
We evaluate the contribution of individual components under a short 2{,}500-step budget (Figure~\ref{fig:ablation}c--d; Appendix~\ref{app:component_ablation}).
For open-vocabulary segmentation, using only the last backbone layer (\emph{Last layer}) costs $6.8$ mAcc, training without puzzles (\emph{No-Puzzles}) $3.0$, removing Gram anchoring $2.8$, and removing cross-view attention (\emph{X-attn.}) $1.9$.
The No-Puzzles variant jointly removes the puzzle-based multi-view formulation and its associated cross-view processing, so this comparison measures their combined contribution rather than isolating the shared-parent supervision alone.
For promptable segmentation, removing LoRA costs $17.1$ mIoU, training only on low-overlap crops (\emph{Hard ovlp.}) $15.0$, removing cross-view attention $4.1$, and removing multi-view prompting (\emph{MVP}) $3.0$.
The Hard-ovlp. result shows that beginning directly with difficult low-overlap views is detrimental under this training budget, rather than isolating the effect of the curriculum schedule itself.
$\mathcal{L}_{\mathrm{warp}}$ and $\mathcal{L}_{\mathrm{cons}}$ have little effect, indicating that these auxiliary consistency losses are not the primary source of the measured performance gains.

\noindent\textbf{Backbone architecture.}
We apply the same recipe to two backbones with different multi-view designs (Table~\ref{tab:backbone_ablation}): VGGT-Omega~\citep{vggtomega}, an offline model where all views attend to each other, and LingBot-Map~\citep{lingbotmap}, a causal streaming model where each frame attends only to past context.
Both use the same head, cached layers, and SA-1B training.
The recipe transfers to LingBot-Map without modification, but VGGT-Omega is better on all metrics (e.g., $+13.7$ mIoU on ScanNet-10).
Since LingBot-Map fits the DINOv3 targets better ($0.816$ \emph{vs.}\ $0.766$ cosine), the gap is unlikely to arise from poorer DINOv3 feature fitting alone and may partly reflect differences in cross-view context, including LingBot-Map's causal attention.
%likely stems from causal attention, which gives early frames only partial scene context.

%--------------------------------------------------------------------------------------
%--------------------------------------------------------------------------------------
%--------------------------------------------------------------------------------------
\section{Conclusion}
% We presented Feature Puzzles, which turns single-image data into synthetic multi-view supervision through overlapping crops with aligned labels. A pretrained VGGT-Omega backbone with selective LoRA supports promptable multi-view segmentation and open-vocabulary 3D features, whose task-specific adaptation requires no real multi-view or video supervision. 
% The model achieves the best mIoU on NVOS, transfers to ScanNet++ and OOD nuScenes-mini tracking, and enables zero-shot open-vocabulary 3D segmentation. With additional in-domain semantic fine-tuning, the same representation surpasses EmbodiedSplat in mIoU, using feed-forward inference without per-scene optimization.
We presented Feature Puzzles, a simple recipe that turns a single-image dataset into synthetic multi-view supervision by cropping a parent image into overlapping sub-images and pulling labels back, so supervision targets are exactly consistent across overlapping affine crop views.
A pretrained VGGT-Omega backbone with layer-selective LoRA feeds two heads---promptable multi-view masks and open-vocabulary 3D features---whose task-specific adaptation requires no real multi-view or video supervision. 
Despite this task-specific supervision requirement, a single forward pass achieves the best mIoU on NVOS, transfers to instance spatial tracking on ScanNet++ and to the out-of-distribution nuScenes-mini setting, and enables zero-shot open-vocabulary 3D segmentation from SA-1B supervision.
With additional in-domain semantic fine-tuning, the same representation further exceeds EmbodiedSplat in mIoU across all evaluated subsets. These results are consistently achieved through efficient feed-forward inference without scene-specific per-scene optimization.

% \david{just to clarify the changes, the conclusion must also preserve the two scope qualifications established throughout the paper: exact consistency refers to the synthetic affine supervision targets, not arbitrary real-camera consistency, and “without video or multi-view supervision” refers to task-specific semantic adaptation, since the geometric backbone itself inherits multi-view pretraining.}

\newpage
\nocite{*}
\bibliography{iclr2027_conference}
\bibliographystyle{iclr2027_conference}

%--------------------------------------------------------------------------------------
%--------------------------------------------------------------------------------------
%--------------------------------------------------------------------------------------
\newpage
\appendix

\section{Feature Puzzle Construction Details}
\label{app:puzzles}

This appendix expands Section~\ref{sec:puzzles}. Throughout, $\pi_n$ denotes the crop-to-parent map, $\ell$ a parent-space label field, and $\ell_n=\mathcal{R}_{\mathcal{Y}}[\ell \circ \pi_n]$ its pull-back to view $n$.

%--------------------------------------------------------------------------------------
\subsection{Teacher Targets: Resolution and Well-Posedness}
\label{app:targets}

The dense teacher target is the frozen teacher evaluated once on the whole parent image, $\ell^{\mathrm{feat}}=\Phi_{\mathrm{DINOv3}}(I_{\downarrow S})$ with $S=2048$, giving a $128 \times 128 \times 768$ parent feature grid. Pull-back to view $n$ is a grid sample at the normalized coordinates $\big(\pi_n^{x}/W,\;\pi_n^{y}/H\big)$ mapped to $[-1,1]^2$, followed by channel-wise L2 normalization, yielding $\mathbf{T}_n \in \mathbb{R}^{768 \times F_r \times F_r}$ with $F_r=128$. Because the normalization uses the \emph{original} parent size $(H,W)$, the aspect-ratio distortion introduced by the square resize to $S$ does not misalign the target.

\paragraph{Resolution.}
A crop of relative scale $s \in [s_{\min},s_{\max}]$ covers $s \cdot 128 \approx 64$--$115$ teacher cells, which are resampled onto the $F_r=128$ target grid. Each view target is therefore a \emph{magnified sub-region of a high-resolution field}, not an upsampled low-resolution one: the student is asked to produce fine-grained spatial structure at $128 \times 128$ from a $16 \times 16$ token grid, which is precisely the signal the RGB-guided decoder of Section~\ref{sec:feature_head} is designed to exploit. A single teacher forward pass is amortized over all $N$ views.

\paragraph{Well-posedness of the objective.}
Vision transformer features are not equivariant under crop-and-resize. Had we defined the target per view as $\mathbf{T}_n=\Phi(x_n)$, two overlapping views would in general disagree on their shared region,
\begin{equation}
    \Phi(x_i)\circ\pi_i^{-1} \;\neq\; \Phi(x_j)\circ\pi_j^{-1}
    \quad \text{on } x_i \cap x_j,
\end{equation}
so the distillation loss $\mathcal{L}_{\mathrm{feat}}$ and the cross-view consistency loss $\mathcal{L}_{\mathrm{warp}}$ would pull the student toward mutually incompatible optima. 
Under the pull-back construction, both losses are consistent with the same parent-derived target $\mathbf{T}_n$, so  
%share the common minimizer $\widehat{\mathbf{F}}_n=\ell^{\mathrm{feat}} \circ \pi_n$, and 
$\mathcal{L}_{\mathrm{warp}}$ acts as a consistent regularizer rather than a competing objective. The same argument applies to the Gram-anchoring term, whose teacher Gram matrix then measures the similarity structure of the \emph{original image} over the region a view covers.

\paragraph{Semantic labels in Stage 2.}
The identical operator transfers the dense dataset-specific parent semantic label map ($-1$ for void) into each view with nearest sampling on the \emph{same} grid, so class labels remain pixel-aligned with the predicted features and the text-aligned cross-entropy of Section~\ref{sec:stage2} is defined on exactly the same lattice as the distillation losses.

%--------------------------------------------------------------------------------------
\subsection{Augmentation, Prompt Derivation, and Validity Filters}
\label{app:prompts}

\paragraph{Augmentation and the pull-back identity.}
The identity $x_n = I \circ \pi_n$, $\ell_n = \ell \circ \pi_n$ is what makes the targets exact, and it is broken by any geometric transform applied to the crop image alone. For the segmentation branch, the loss requires image-target alignment within each view, so geometric, photometric, and occlusion augmentations are applied jointly and synchronously to $(x_n,\mathbf{G}_n,\mathrm{prompt}_n)$. For the feature branch, whose target lives in parent space, geometric and occlusion augmentation are disabled and only photometric jitter is retained, preserving pixel-exact student--teacher correspondence for the affine crop construction.

\paragraph{Prompts as noisy observations of the target.}
Prompts are derived \emph{after} augmentation from the crop-space target $\mathbf{G}_{n,m}$, so they are expressed in final crop pixel coordinates and are pixel-aligned with the image by construction. A prompt is always a stochastic degradation $\rho(\mathbf{G}_{n,m})$ of the target, never the target itself:

\begin{itemize}
\item \textbf{Mask.} A random morphological dilation or erosion with an elliptical kernel $k \le \min(9,\lfloor 0.25\,d \rfloor)$, where $d$ is the object's shorter side, guarded so that small masks cannot vanish, plus boundary noise.
\item \textbf{Box.} The tight bounding box of a morphologically perturbed copy of the mask, with $\pm 2\%$ side jitter.
\item \textbf{Point.} One to three positives sampled inside the mask and up to $P-n_{\mathrm{pos}}$ negatives sampled outside it.
\end{itemize}

The label $\mathbf{G}_{n,m}$ is left untouched. Treating the prompt as a noisy observation rather than as a copy of the label trains prompt robustness without injecting label noise.
For each instance, a primary prompt view is sampled uniformly from its visible views. With probability $p_{\mathrm{mv}}=0.5$, additional visible views are also prompted, with each prompt independently derived from the corresponding view target.
% For each instance a primary prompt view is drawn uniformly among the views where it is visible, and with probability $p_{\mathrm{mv}}=0.5$, prompts are also provided in a random non-empty subset of the remaining visible views (multi-view prompting).

% \david{if the implementation actually re-derives each additional prompt from that view's target, say so explicitly. For example, "additional prompts are independently derived from the corresponding view targets.”} \jiahao{fix}

\paragraph{Validity criteria.}
Three filters keep the transferred labels informative.

\begin{enumerate}
\item \emph{Annotation quality}, applied in parent space before transfer: area $\ge 1024$\,px$^2$, SAM predicted IoU $\ge 0.85$, stability score $\ge 0.9$, and at most $120$ annotations per parent. The parent-space area threshold also helps ensure that instances remain resolvable after the crop-to-$R$ resampling.
\item \emph{Cross-view support}: an instance is retained only if it is non-empty in at least $\kappa=3$ of the $N$ views, since instances visible in fewer views contribute mostly empty channels and little cross-view signal; at most $M=16$ instances are kept per sample.
\item \emph{Foreground coverage} of the boxes themselves: a candidate box is accepted only if the mean of the union-of-annotations mask inside it is at least $\tau_{\mathrm{fg}}=0.5$, which prevents crops that fall on unannotated background.
\end{enumerate}

\paragraph{Truncated and absent instances.}
An instance that is truncated or absent in view $n$ is \emph{not} re-fitted: its visible part is retained, and $\mathbf{G}_{n,m}=\mathbf{0}$ when the box misses it entirely.

%--------------------------------------------------------------------------------------
\subsection{Detailed Comparison with Image-to-Clips}
\label{app:i2c}

Image-to-Clips~\citep{ma2025puzzles} targets 3D reconstruction, so each synthesized patch must carry geometric labels. It assigns a per-patch camera by solving a PnP problem with RANSAC over the back-projected depth, applies a centroid-based random rotation, re-renders the point cloud, and then \emph{validates} the result with front-surface-coverage and view-frustum tests, discarding views whose occlusion holes are too large. The resulting labels are estimates, with residual error contributed by depth, by pose fitting, and by rendering.

Our Feature Puzzles views include both 3D augmentations and in-plane crop-and-resize transformations. Since the supervision is appearance-based (masks, dense features, and semantic classes) rather than geometric, both constructions preserve consistent parent-derived semantics, while the affine crop views additionally provide exact view-to-parent mappings. Table~\ref{tab:i2c_comparison} summarizes the two constructions term by term.

\begin{table}[t]
\centering
\small
\caption{Term-by-term comparison between the Image-to-Clips augmentation of Puzzles~\citep{ma2025puzzles} and the Feature Puzzles construction.}
\label{tab:i2c_comparison}
\begin{tabular}{lll}
\toprule
 & Image-to-Clips (Puzzles) & Feature Puzzles (ours) \\
\midrule
Parent annotation & RGB-D (GT or predicted depth) & \makecell[l]{Instance masks / semantic map /\\frozen teacher field} \\
\addlinespace
View-to-parent map & \makecell[l]{Fixed $\mathbf{K}$ + extrinsics from\\PnP-RANSAC + centroid rotation} & Closed-form affine $\pi_n$ \\
\addlinespace
Label construction & \makecell[l]{Per-patch depth and pose,\\re-rendered} & Pull-back $\ell \circ \pi_n$ \\
\addlinespace
Error sources & \makecell[l]{Depth, PnP residual,\\rendering holes} & Resampling only \\
\addlinespace
Validity handling & \makecell[l]{FrontCov / ImgCov tests,\\reject or hole-fill} & \makecell[l]{Foreground coverage +\\cross-view support} \\
\addlinespace
Cross-view agreement & Approximate & Exact by construction \\
\bottomrule
\end{tabular}
\end{table}

%Viewpoint diversity, which Image-to-Clips obtains from simulated 6-DoF motion, is recovered in our construction through the scale range $[s_{\min},s_{\max}]$, the overlap curriculum, and photometric augmentation.
While Image-to-Clips introduces viewpoint diversity through simulated 6-DoF motion, our affine construction instead introduces crop, scale, and appearance variation through the scale range $[s_{\min},s_{\max}]$, the overlap curriculum, and photometric augmentation. It does not simulate parallax or disocclusion.

%--------------------------------------------------------------------------------------
\subsection{Detailed Comparison with VGGT-Segmentor}
\label{app:vggts}

VGGT-Segmentor~\citep{gao2026vggtsegmentor} targets ego--exo cross-view segmentation. Given a synchronized image pair and a source-view mask, it fuses the mask into VGGT features, projects points sampled from the source mask into the target view with the VGGT track head, and refines the resulting target mask. Its main model is trained on paired Ego-Exo4D masks, and a correspondence-free variant is pre-trained with single-image self-supervision on a subset of SA-1B. Both variants remain pairwise at training and inference.

Feature Puzzles differs in the unit of reasoning. Pairwise methods match every target independently to a reference, so a target seen from a very different viewpoint cannot benefit from intermediate views unless pairwise associations or tracking chains are built explicitly. 
For large unordered view sets, this can require repeated pairwise processing and explicit association across views.
%This becomes the dominant cost for hundreds of randomly ordered observations. 
Feature Puzzles trains on $N$-view puzzle sequences and predicts masks for all views jointly: cross-view attention allows information to be exchanged across all views, including through intermediate observations, 
%lets intermediate observations bridge large viewpoint changes, 
prompts from any subset of views are fused into shared object tokens, and consistency is supervised at the sequence level through targets in every view. No point correspondence, pair selection, tracking, or view ordering is needed. Table~\ref{tab:vggts_comparison} summarizes these differences.

\begin{table}[t]
\centering
\caption{Detailed comparison between VGGT-Segmentor~\citep{gao2026vggtsegmentor} and Feature Puzzles.}
\label{tab:vggts_comparison}
\small
\begin{tabular}{lcc}
\toprule
\textbf{Aspect} & \textbf{VGGT-Segmentor} & \textbf{Feature Puzzles} \\
\midrule
Training unit
& Image pair
& Multi-view sequence \\

Inference
& Source--target pairs
& All views jointly \\

Object modeling
& Pairwise correspondence
& Joint multi-view representation \\

Prompt propagation
& Source $\rightarrow$ target
& Arbitrary view(s) $\rightarrow$ all views \\

Explicit correspondence
& VGGT point tracking
& Not required \\

Intermediate-view context
& Pairwise only %Not exploited
& Jointly exploited \\

Unordered view set
& Processed via source--target pairs %Requires pair selection/association
& No explicit association \\

Cross-view consistency
& Pair-level
& Sequence-level \\

Prompt conditioning %Multi-prompt fusion
& Source-view prompt %Limited
& Multiple prompted views supported %Supported 
\\

\bottomrule
\end{tabular}
\end{table}

% \david{Please check if you agree with the changes in the Table. "Source-view prompt" versus "Multiple prompted views supported" states the architectural difference directly.}

%--------------------------------------------------------------------------------------
%--------------------------------------------------------------------------------------
%--------------------------------------------------------------------------------------
\section{Architecture and Training Details}
\label{app:arch}

%--------------------------------------------------------------------------------------
\subsection{Backbone Adaptation}
\label{app:backbone}

The pretrained backbone jointly processes the $N$ crops and produces geometry-aware patch tokens. Because intermediate and final layers carry complementary local, semantic, and geometric information, we cache tokens $\mathbf{T}_{\ell}\in\mathbb{R}^{B \times N \times H_p \times W_p \times 2048}$ from four blocks $\mathcal{L}=\{4,11,17,23\}$. The fused representation $\mathbf{F}$ is projected to a working dimension of $768$ inside each task head. Low-rank adapters are inserted only in the late blocks (18--23), so blocks 0--17 stay untouched and still serve the original depth and camera heads. Keeping most backbone parameters frozen helps preserve the representation learned during multi-view pre-training.

Each cached layer receives a lightweight low-rank residual and the layers are combined by a softmax over learned logits $a_\ell$:
\begin{equation}
\label{eq:fusion}
    \widetilde{\mathbf{T}}_{\ell}
    = \operatorname{norm}(\mathbf{T}_{\ell})
    + (\mathbf{T}_{\ell}\mathbf{A}_{\ell})\mathbf{B}_{\ell},
    \qquad
    \mathbf{F} =
    \sum_{\ell \in \mathcal{L}}
    \frac{\exp(a_\ell)}{\sum_{j \in \mathcal{L}}\exp(a_j)}
    \widetilde{\mathbf{T}}_{\ell},
\end{equation}
where $\mathbf{A}_{\ell}\in\mathbb{R}^{2048\times r}$ is the LoRA down-projection and $\mathbf{B}_{\ell}\in\mathbb{R}^{r\times 2048}$ the up-projection of rank $r$, so $(\mathbf{T}_{\ell}\mathbf{A}_{\ell})\mathbf{B}_{\ell}$ is a low-rank update to the normalized tokens; the fusion is a softmax-weighted sum of the adapted tokens across the four cached layers.

Adapter rank increases with depth: starting from a base rank of $16$, it is multiplied by two for middle blocks and by four for late blocks, and inter-frame blocks receive an additional factor of two. We scale $\alpha$ with rank so that $\alpha/r$ remains constant. The adapter output projection and the layer-fusion residual branches are initialized to zero, and the fusion logits uniformly, so the LoRA-adapted backbone initially reproduces the pre-trained backbone, while the fusion begins as an equal-weight average. This makes the module plug-and-play and permits two-pass inference: adapters off for reconstruction, on for segmentation.

%--------------------------------------------------------------------------------------
\subsection{Promptable Segmentation Head}
\label{app:promptable}

\paragraph{Prompt encoding and token layout.}
The prompt encoder produces a dense coverage map $\mathbf{C}\in\mathbb{R}^{B\times N\times M\times H_p\times W_p}$, a prompt-presence indicator, and sparse prompt tokens. Conditioned tokens are projected to a working dimension of $768$. Cross-view self-attention operates over the $N$ view tokens at each spatial patch, followed by four SAM-style two-way attention blocks~\cite{ravi2024sam2}. Four object tokens per view and instance are initialized by aggregating prompt-covered image tokens across all prompted views; sparse prompt tokens are appended to the query sequence, with padded prompts removed by a key-padding mask. Image and object tokens attend to one another in both directions. When depth and camera poses are available, a 3D positional encoding is added to the image tokens. The weighted pixel decoder projects and fuses decoder features and upsamples them to $\mathbf{D}\in\mathbb{R}^{B\times N\times C\times H\times W}$, and the mask object token predicts the dynamic kernel $\mathbf{k}\in\mathbb{R}^{B\times N\times M\times C}$.

\paragraph{Loss terms and design.}
We set $(\lambda_{\mathrm{bce}},\lambda_{\mathrm{dice}},
\lambda_{\mathrm{iou}},
\lambda_{\mathrm{cons}},\lambda_{\mathrm{pt}})
=(5,5,1,0.01,1)$. 
The design follows one principle: because pull-back places a target $\mathbf{G}_{n,m}$ in \emph{every} view rather than only the prompted one, mask supervision can be evaluated across all $N$ views, directly encouraging cross-view agreement even where no prompt is provided. Additional terms regularize confidence, object tokens, and prompted point locations.
%so cross-view consistency is supervised directly instead of being encouraged indirectly.

\begin{itemize}
\item $\mathcal{L}_{\mathrm{bce}}$, $\mathcal{L}_{\mathrm{dice}}$: pixel-wise binary cross-entropy with logits and Dice loss between $\sigma(\mathbf{Y}_{n,m})$ and $\mathbf{G}_{n,m}$, summed over all views. Since a target exists in views that received no prompt, a prediction correct only in the prompted view is penalized, so the mask term alone already drives cross-view agreement. Dice complements the pixel-wise term on small instances, where background dominates.
\item $\mathcal{L}_{\mathrm{iou}}$: regression of the auxiliary head onto the measured overlap between $\sigma(\mathbf{Y}_{n,m})$ and $\mathbf{G}_{n,m}$, giving a per-view confidence used to rank masks at inference.
\item $\mathcal{L}_{\mathrm{cons}}$: cross-view variance of the mask object token around its centroid, $\tfrac{1}{N}\sum_n \|\mathbf{t}_{n,m}-\bar{\mathbf{t}}_m\|_2^2$ with $\bar{\mathbf{t}}_m=\tfrac{1}{N}\sum_n \mathbf{t}_{n,m}$. 
The $N$ tokens describe the same physical object, so they are encouraged to remain close in token space; this regularizes the \emph{representation} rather than the output. Its weight is kept small ($0.01$) because collapsing the tokens too hard removes the per-view information the dynamic kernel needs.
\item $\mathcal{L}_{\mathrm{pt}}$: bilinearly sampled binary cross-entropy at the prompt point locations, raising the probability at positives and lowering it at negatives. 
This provides direct supervision at the prompted locations, encouraging positive clicks to activate and negative clicks to suppress the predicted region.
%Without this term negative clicks are merely encoded and then ignored; the term is what makes them actively suppress the predicted region.
\end{itemize}

%--------------------------------------------------------------------------------------
\subsection{Open-Vocabulary Head}
\label{app:openvocab}

\paragraph{Visual and text teachers.}
During training the complete parent image is square-resized to $2048\times2048$ and processed by a frozen DINOv3~\citep{simeoni2025dinov3} ViT-B/16~\citep{dosovitskiy2021vit}, whose dense patch tokens form the $128\times128$ parent feature map; per-crop targets are obtained by the pull-back of Appendix~\ref{app:targets}. At inference, class names are inserted into an ImageNet-style prompt ensemble and encoded by a frozen CLIP~\citep{radford2021clip} ViT-B/16 text tower, and a frozen Talk2DINO~\citep{barsellotti2025talk2dino} MLP projects the $512$-dimensional CLIP embeddings into the $768$-dimensional DINOv3 space. Both image and text embeddings are L2-normalized, allowing classification by cosine similarity.

\paragraph{Stage-1 losses: distillation, warping, and Gram anchoring.}
The weights of $\mathcal{L}_{\mathrm{feat}}$, $\mathcal{L}_{\mathrm{warp}}$, and $\mathcal{L}_{\mathrm{gram}}$ are $1.0$, $0.1$, and $0.5$. The distillation term $\mathcal{L}_{\mathrm{feat}}$ is the mean pixel-wise cosine distance between the student $\widehat{\mathbf{F}}_n$ and the pulled-back teacher $\mathbf{T}_n$; on its own it supervises each view in isolation. The other two terms (defined in Section~\ref{sec:feature_head}) exploit the shared-parent correspondence provided by the puzzle construction.
%are where the puzzle construction pays off, and both exist only because $\pi_n$ is exact.

\emph{Cross-view warping.} For an overlapping pair $(i,j)$ the composite map $\pi_i^{-1}\!\circ\pi_j$ carries crop $j$'s grid into crop $i$'s, so the student prediction of view $i$ can be transported into view $j$ and compared there. 
The stop-gradient $\mathrm{sg}[\cdot]$ prevents gradients from flowing through the target branch,  and the valid set $\Omega_{ij}=\{q \in \Omega_R : \pi_j(q) \in x_i\}$ is read directly off the sub-image boxes in parent coordinates---no depth test, no visibility reasoning, no photometric matching. At most $12$ pairs are drawn per step. 
%This term is what turns cross-view consistency into an explicit training signal rather than a property hoped for at inference, and it is precisely what a single-view objective cannot express.
This provides an explicit pairwise cross-view consistency signal that is unavailable to a purely single-view objective.

\emph{Gram anchoring.} For $K=512$ pixels sampled per step we form the $K\times K$ pairwise cosine-similarity matrices of student and teacher and match them. Because every view's teacher is the \emph{parent} feature field restricted to that view, $\mathbf{S}_{T}\mathbf{S}_{T}^\top$ encodes the similarity structure of the original image over the region the view covers. 
The Gram objective is designed to preserve feature dissimilarity across object boundaries, helping counteract spatial smoothing when predicting at $128\times128$ from a $16\times16$ token grid.
Because views at different scales inherit their target from the same parent field, it also encourages them to share a common similarity structure, complementing the pixel-level agreement  enforced by $\mathcal{L}_{\mathrm{warp}}$.

\paragraph{Stage-2 class weighting.}
We use effective inverse-frequency class weights with $\beta=0.999$ and focal modulation $(1-p_t)^\gamma$ with $\gamma=2$. Auxiliary text-classification heads cover the ScanNet20, ScanNet200, and ScanNet++ vocabularies while sharing the same dense feature map.

%--------------------------------------------------------------------------------------
\subsection{Implementation Details}
\label{app:impl}

\textit{Common training setup.}
Both tasks use cached VGGT-Omega blocks $\{4,11,17,23\}$, 2048-dimensional backbone tokens, a working dimension of 768, bfloat16 mixed precision, AdamW, and cosine learning-rate decay with linear warm-up. Training is distributed over eight A800 GPUs.

\textit{Promptable segmentation.}
We train with $N=6$ views, four object tokens, four two-way attention layers, and eight attention heads. The probability of placing prompts in multiple views is $0.5$, and the supported modalities are masks, boxes, and points with at most six points per prompt. Low-rank adaptation starts at block 18 with base rank \(r=16\) and scale \(\alpha=16\).

\textit{Open-vocabulary segmentation.}
For open-vocabulary pre-training, we use six $256\times256$ crops, predict $128\times128$ features, sample 512 pixels for the Gram objective, and use at most 12 view pairs for the warp objective. ScanNet fine-tuning uses a learning rate of $5\times10^{-5}$. At inference, frames are sampled with stride 10 and fused using $0.02$\,m voxels.

\textit{Inference details.}
The architecture supports prompts from one or multiple views, which are encoded jointly, and masks for all views are returned in one forward pass, with no scene-specific parameters optimized. The same backbone still serves depth and camera reconstruction when the segmentation adapters are disabled.

\begin{table}[t!]
\centering
\caption{Component ablations for open-vocabulary 3D segmentation on EmbodiedSplat (Stage-1 only, 2{,}500-step budget, no in-domain fine-tuning). $\Delta$ is the ScanNet 10-class mAcc change relative to the full model. Bold = best, underline = second best.}
\label{tab:component_ablation}
\resizebox{\textwidth}{!}{%
\begin{tabular}{lccccccccc}
\toprule
\multirow{2}{*}{Variant} & \multicolumn{6}{c}{ScanNet} & \multicolumn{2}{c}{ScanNet200} & \multirow{3}{*}{$\Delta$mAcc$_{10}$} \\
\cmidrule(lr){2-7}\cmidrule(lr){8-9}
 & \multicolumn{2}{c}{10 cls} & \multicolumn{2}{c}{15 cls} & \multicolumn{2}{c}{19 cls} & \multicolumn{2}{c}{70 cls} & \\
\cmidrule(lr){2-3}\cmidrule(lr){4-5}\cmidrule(lr){6-7}\cmidrule(lr){8-9}
 & mIoU & mAcc & mIoU & mAcc & mIoU & mAcc & mIoU & mAcc & \\
\midrule
Full model & \textbf{15.70} & \textbf{29.69} & \textbf{10.01} & \textbf{19.95} & \textbf{8.46} & \textbf{18.73} & 3.32 & \underline{7.59} & -- \\
\midrule
w/o $\mathcal{L}_{\mathrm{warp}}$ & \underline{15.60} & \underline{29.49} & \underline{9.86} & \underline{19.87} & \underline{8.22} & \textbf{18.73} & \textbf{3.64} & \textbf{7.87} & $-0.20$ \\
w/o cross-view attention & 14.07 & 27.77 & 8.89 & 18.57 & 7.35 & 17.48 & \underline{3.37} & 7.42 & $-1.92$ \\
w/o $\mathcal{L}_{\mathrm{gram}}$ & 12.85 & 26.91 & 8.10 & 18.06 & 7.05 & \underline{17.60} & 2.74 & 6.48 & $-2.78$ \\
No-Puzzles (single-view) & 13.32 & 26.70 & 8.43 & 17.89 & 7.35 & 17.28 & 3.21 & 7.37 & $-2.99$ \\
Last layer only (block 23) & 8.93 & 22.89 & 5.76 & 15.38 & 4.85 & 14.78 & 2.05 & 4.84 & $-6.80$ \\
\bottomrule
\end{tabular}%
}
\end{table}

\begin{table}[t!]
\centering
\caption{Component ablations for promptable multi-view segmentation on SceneNN-MV (GT-prompted, 6 views, 2{,}500-step budget). ``Vis'' metrics are averaged over views where the object is visible; ``Obj'' is object-level mIoU. $\Delta$ is the Vis mIoU change relative to the full model. Bold = best, underline = second best.}
\label{tab:component_ablation_seg}
\small
\begin{tabular}{lccccc}
\toprule
Variant & Vis mIoU & Obj mIoU & Vis Prec. & Vis Rec. & $\Delta$Vis mIoU \\
\midrule
Full model (LoRA + multi-view prompting) & \textbf{55.19} & \textbf{56.80} & \underline{65.35} & \underline{76.79} & -- \\
\midrule
w/o $\mathcal{L}_{\mathrm{cons}}$ & \underline{55.15} & \underline{56.36} & 65.11 & 76.74 & $-0.04$ \\
w/o multi-view prompting & 52.17 & 52.01 & 60.53 & \textbf{77.97} & $-3.02$ \\
w/o cross-view attention & 51.12 & 54.66 & \textbf{67.80} & 67.37 & $-4.07$ \\
Hard overlap $[0.10,0.50]$ & 40.23 & 41.39 & 51.20 & 61.75 & $-14.96$ \\
w/o LoRA & 38.11 & 36.40 & 47.10 & 64.37 & $-17.08$ \\
Independent views (no shared parent) & 6.49 & 8.21 & 7.84 & 5.62 & $-48.70$ \\
\bottomrule
\end{tabular}
\end{table}

%--------------------------------------------------------------------------------------
%--------------------------------------------------------------------------------------
%--------------------------------------------------------------------------------------
\section{Additional Ablations}
\label{app:component_ablation}

\paragraph{Open-vocabulary 3D segmentation.} All arms share the Stage-1 open-vocabulary setup: frozen VGGT-Omega, SA-1B Feature Puzzles with $N=6$ views, $256\times256$ crops, and loss weights $(\lambda_{\mathrm{feat}},\lambda_{\mathrm{warp}},\lambda_{\mathrm{gram}})=(1.0,0.1,0.5)$. Each arm is trained for a shortened budget of 2{,}500 steps (learning rate $2\times10^{-4}$, no warm-up, batch 12 per GPU on seven A800 GPUs). 
Ablations modify one component unless otherwise noted. \emph{No-Puzzles} feeds each crop as an independent single-view sequence, which also disables head cross-view attention and $\mathcal{L}_{\mathrm{warp}}$. \emph{Last layer only} caches block 23 instead of $\{4,11,17,23\}$. Evaluation uses the EmbodiedSplat protocol (stride 10, at most 400 frames, $0.02$\,m voxels) without ScanNet fine-tuning, so absolute values are not comparable to Table~\ref{tab:embodiedsplat_acc}. Table~\ref{tab:component_ablation} reports the results.

% \david{Regarding Table 8, I have a question about the table row name "w/o Feature Puzzles." If that row is actually the independently sampled-view experiment described here, something like "Independent views" or "Independent views (no shared parent)" would be clearer and more faithful to what was tested.}

\paragraph{Promptable segmentation.} All arms share the promptable setup of Appendix~\ref{app:impl} ($N=6$ views, four object tokens, four two-way layers, mask/box/point prompts) and are trained from the pretrained VGGT-Omega weights for 2{,}500 steps (learning rate $6\times10^{-5}$, no warm-up, batch 6 per GPU on eight A800 GPUs). \emph{w/o LoRA} keeps the backbone fully frozen. \emph{w/o multi-view prompting} sets $p_{\mathrm{mv}}=0$. \emph{Hard overlap} trains directly on the end state of the overlap curriculum, $[0.10,0.50]$, instead of the default $[0.70,0.95]$; the one-epoch budget is too short to anneal the schedule. Evaluation uses GT-prompted 6-view SceneNN-MV (12 scenes, 1{,}005 objects) without refinement. For reference, the fully trained model reaches $70.26$ visible-view mIoU. A literal ``w/o Feature Puzzles'' (bottom of Table~\ref{tab:component_ablation_seg}) setting is not directly trainable because the puzzle construction defines the cross-view supervision. A cleaner ablation is therefore to keep the same $N$-view architecture and training setup but replace shared-parent crops with independently sampled views, removing the shared-parent correspondence while retaining the same multi-view architecture and training budget. Table~\ref{tab:component_ablation_seg} reports the results.

\begin{figure}[t]
    \centering
    \includegraphics[width=\linewidth]{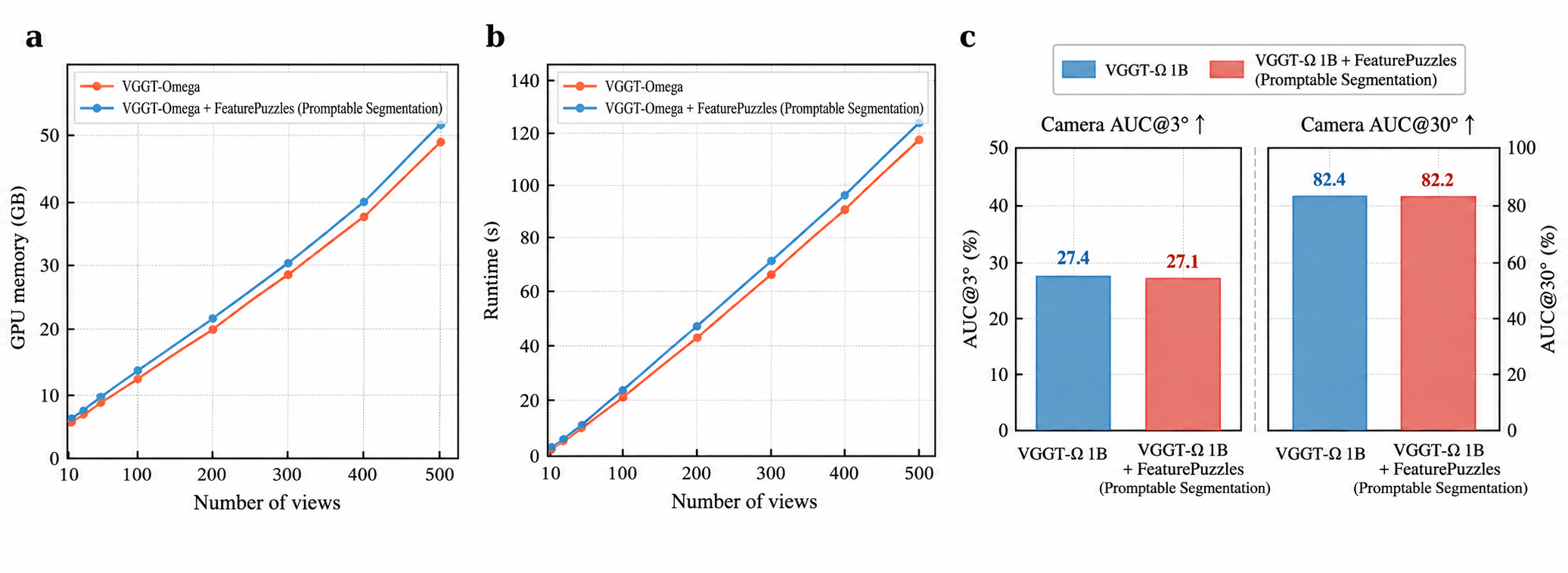}
    \vspace{-0.8cm}
    \caption{\textbf{Efficiency and camera-pose preservation.}
(a--b) GPU memory and runtime of VGGT-$\Omega$ with and without Feature Puzzles as the number of views increases. 
(c) Camera pose accuracy on 7-Scenes, showing that Feature Puzzles largely preserves the original camera-pose performance.}
    \label{fig:comp_with_default}
\end{figure}

% \noindent
% \begin{minipage}[t]{0.47\linewidth}
% \vspace{0pt}
% \textbf{Multi-view prompting.}
% Performance consistently improves as the number of prompted views increases from one to five. mIoU rises from 43.35\% to 64.98\%, while T-SR improves from 74.50\% to 86.77\%, showing that prompts from multiple views provide complementary cues for joint segmentation. This directly supports the benefit of multi-view prompting beyond the single-view setting. When all 10 views are prompted, however, both metrics decrease substantially, suggesting that excessive prompts may introduce redundant or conflicting information. 
% \end{minipage}
% \hfill
% \begin{minipage}[t]{0.49\linewidth}
% \vspace{0pt}
% \centering
% \small
% \begin{tabular}{c|cc}
% \toprule
% \textbf{\# Prompts} & \textbf{mIoU} & \textbf{T-SR} \\
% \midrule
% 1  & 43.35 & 74.50 \\
% 2  & 49.55 & 81.12 \\
% 3  & 54.98 & 84.32 \\
% 5  & \textbf{64.98} & \textbf{86.77} \\
% 10 & 39.98 & 62.18 \\
% \bottomrule
% \end{tabular}
% \captionof{table}{\textbf{Effect of multi-view prompting on ScanNet.} Performance with different numbers of prompted views.}
% \label{tab:multi_prompt}
% \end{minipage}

\paragraph{Efficiency and Camera-Pose Preservation.} As shown in Figure~\ref{fig:comp_with_default}, Feature Puzzles introduces only marginal computational overhead over the original VGGT-$\Omega$. GPU memory and runtime scale similarly with the number of input views, as the semantic head reuses the existing multi-view backbone features without introducing an additional image encoder. Meanwhile, camera-pose performance is largely preserved: on 7-Scenes, Camera AUC@3$^\circ$ changes only from 27.4 to 27.1, and AUC@30$^\circ$ from 82.4 to 82.2.

%--------------------------------------------------------------------------------------
%--------------------------------------------------------------------------------------
%--------------------------------------------------------------------------------------
\section{Feature Puzzles Label}
\label{app:feature_puzzles_label}
\paragraph{Exact Cross-View Label Correspondence.} 
We define \textbf{exact cross-view label correspondence} as pixel-to-label alignment within the synthetic-view construction, rather than exact correspondence to the true 3D scene. For affine crops, images and labels share the same known transformation. For 3D-augmented views, rendered RGB and labels use the same estimated depth and virtual-camera projection; depth errors may affect geometric realism but not their mutual pixel alignment under that projection. Since all views inherit supervision from the same parent label field, shared visible regions remain label-consistent, while disoccluded regions are masked out from the loss.

%--------------------------------------------------------------------------------------
%--------------------------------------------------------------------------------------
%--------------------------------------------------------------------------------------
\section{Qualitative Results}
\label{app:vis_promptable_seg}
\begin{figure*}[t]
\centering
\includegraphics[width=\textwidth]{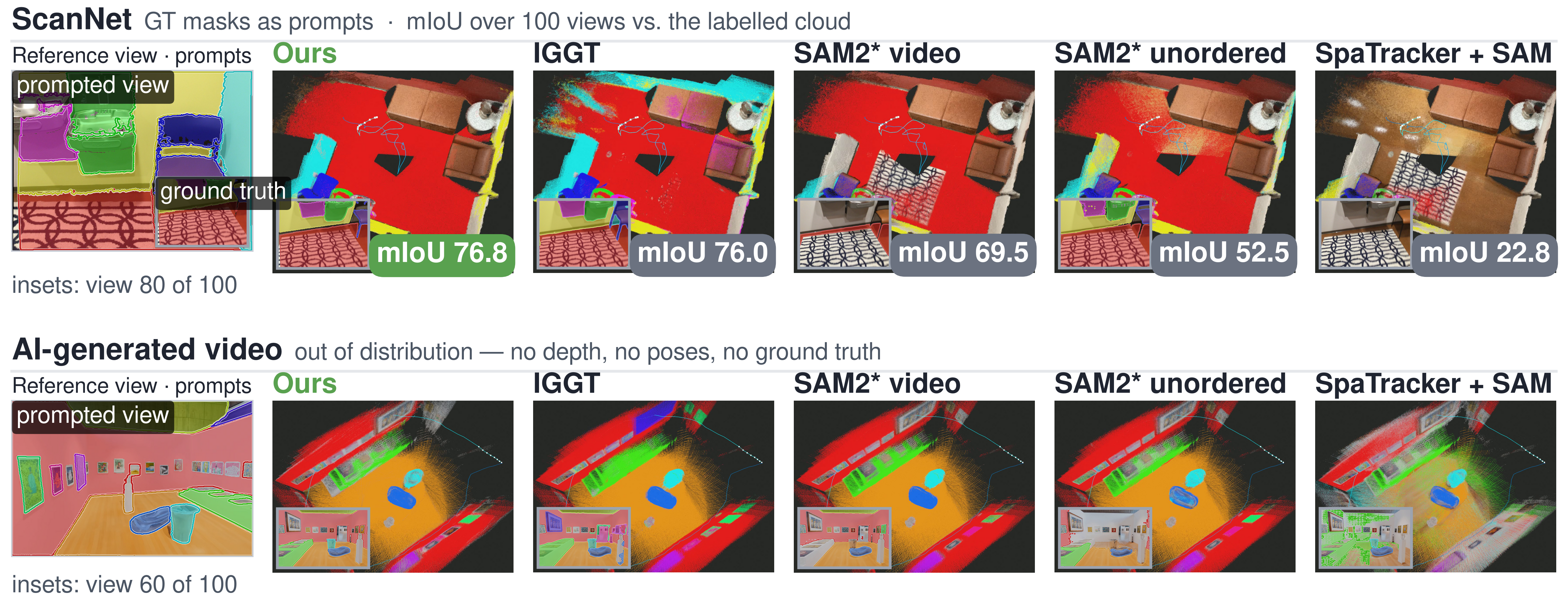}
\vspace{-0.6cm}
\caption{\textbf{Promptable multi-view segmentation.} 
Results on ScanNet and an out-of-distribution AI-generated video, showing consistent cross-view mask propagation from sparse input prompts.
% \david{I would move this to supplement. Useful visual evidence, but quantitative results already in Table 1/2} \jiahao{Agree.}
}
\label{fig:promptable_seg}
\end{figure*}

\paragraph{Promptable multi-view segmentation.}
Figure~\ref{fig:promptable_seg} provides qualitative results complementary to Tables~\ref{tab:nvos_spinerf} and~\ref{tab:tracking_merged}. On ScanNet and an out-of-distribution AI-generated video, Feature Puzzles propagates prompts from a reference view to a large set of views in a single forward pass. Compared with IGGT, SAM2, and SpaTracker+SAM, the predictions show greater consistency across viewpoint changes and better preserve object identity over long sequences.

\begin{figure*}[t]
\centering
\includegraphics[width=\textwidth]{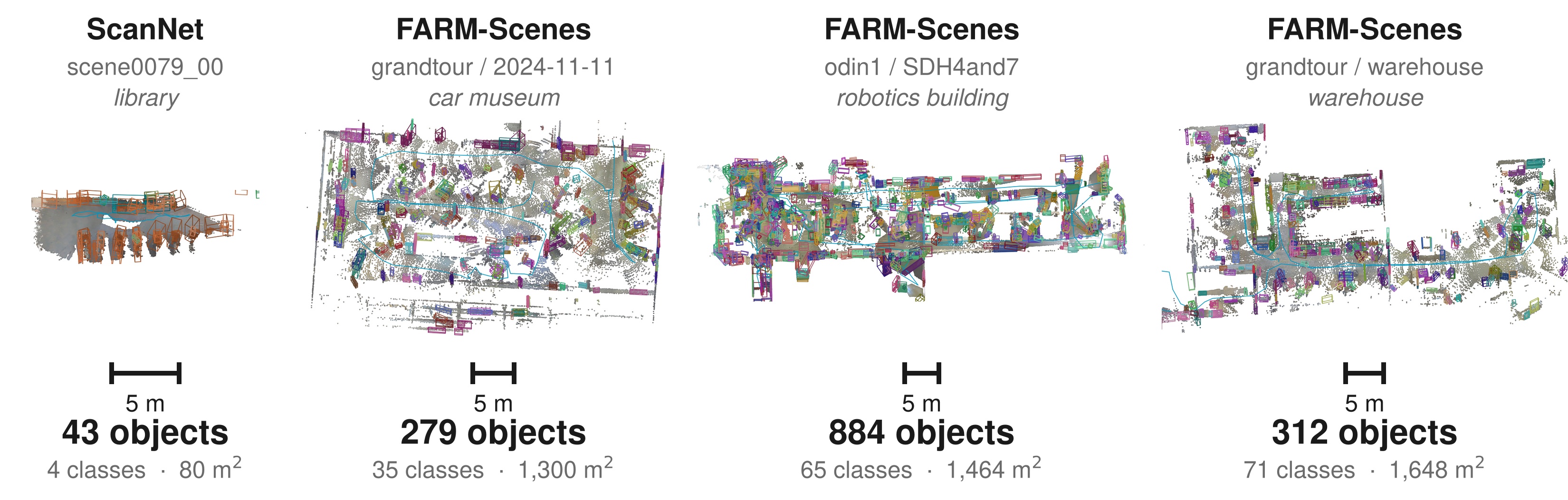}
\vspace{-0.5cm}
\caption{\textbf{Open-vocabulary 3D segmentation.} 
Zero-shot evaluation on ScanNet and FARM-Scenes, illustrating coherent semantic reconstruction across diverse large-scale environments.
% \david{I would move this to suplement. Table 3 carries the scientific comparison} \jiahao{Agree.}
}
\label{fig:open_vocab_seg}
\end{figure*}

\paragraph{Open-vocabulary 3D segmentation.}
Figure~\ref{fig:open_vocab_seg} visualizes zero-shot open-vocabulary 3D segmentation on ScanNet and FARM-Scenes. The learned semantic features generalize across indoor scenes and substantially larger, more diverse environments, including a car museum, robotics building, and warehouse. These results illustrate that the feed-forward representation can produce coherent 3D semantic predictions across scenes with large differences in scale, layout, and object diversity.

\end{document}

%% file: math_commands.tex
\usepackage{amsmath,amsfonts,bm}

\def\eqref#1{equation~\ref{#1}}
\def\1{\bm{1}}

\def\re{{\textnormal{e}}}

\DeclareMathAlphabet{\mathsfit}{\encodingdefault}{\sfdefault}{m}{sl}
\SetMathAlphabet{\mathsfit}{bold}{\encodingdefault}{\sfdefault}{bx}{n}